\documentclass[lettersize,journal]{IEEEtran}
\usepackage[pdftex]{graphicx}
\usepackage{pgfplots}
\usepackage{xcolor,colortbl}
\usepackage{tikz,booktabs}
\usepackage{amsmath}
\usepackage{amsmath}
\usepackage{amsthm}
\usepackage{amssymb}
\usepackage{mathrsfs}
\usepackage{cite} 
\usepackage{ulem}
\usepackage{graphicx}
\usepackage{multirow}
\usepackage{array}
\usepackage[caption=false,font=normalsize,labelfont=sf,textfont=sf]{subfig}
\usepackage{makecell}
\usepackage{tikz}

\usepackage[framemethod=TikZ]{mdframed}
\usepackage{verbatim}
\usepackage{smartdiagram}
\usepackage{cite,algpseudocode,algorithm,multirow,amssymb,latexsym,amsfonts,tikz,adjustbox,pgfplots,pgf-umlsd,enumerate,enumitem,epstopdf,graphicx,changepage}
\usepackage{color}
\usetikzlibrary{matrix,chains,positioning,decorations.pathreplacing,arrows,positioning,fit,shapes}
\usetikzlibrary{arrows,calc,fit,decorations.text}
\usetikzlibrary{shapes,arrows}
\usetikzlibrary{fit}

\usepackage{amsmath}
\usepackage{pifont}

\newcommand{\xmark}{\text{\ding{55}}}
\newcommand{\argmax}{\operatornamewithlimits{argmax}}

\begin{document}
\title{Motion Comfort Optimization for Autonomous Vehicles: Concepts, Methods, and Techniques}


\author{Mohammed~Aledhari,~\IEEEmembership{Senior Member,~IEEE,}
Mohamed Rahouti,~\IEEEmembership{Member,~IEEE,}
Junaid Qadir,~\IEEEmembership{Senior Member,~IEEE,}
Basheer Qolomany,~\IEEEmembership{Member,~IEEE,}
Mohsen Guizani,~\IEEEmembership{Fellow,~IEEE,}  
Ala Al-Fuqaha,~\IEEEmembership{Senior Member,~IEEE}
\thanks{M. Aledhari is with the Department of Data Science, University of North Texas, Denton, TX, 76207 USA. Email: mohammed.aledhari@unt.edu}
\thanks{M. Rahouti is with the Department of Computer and Information Science, Fordham University, Bronx, NY, 10458 USA. Email: mrahouti@fordham.edu}
\thanks{J. Qadir is with the Qatar University, Doha, Qatar. Email: jqadir@qu.edu.qa}
\thanks{B. Qolomany is with the Department of Cyber Systems,  
University of Nebraska at Kearney, Kearney, NE, 68849 USA. Email: qolomanyb@unk.edu}
\thanks{M. Guizani is with Mohamed Bin Zayed University of Artificial Intelligence (MBZUAI), Abu Dhabi, UAE. Email: mguizani@ieee.org}
\thanks{A. Al-Fuqaha is with the Information and Computing Technology (ICT) Division, College of Science and Engineering (CSE), Hamad Bin Khalifa University, Doha, Qatar. Email: aalfuqaha@hbku.edu.qa}
\thanks{Manuscript received xx, 2023; revised xx, 2023.}}

\markboth{IEEE Internet of Things Journal,~Vol.~xx, No.~xx, xx~2023}%
{Aledhari \MakeLowercase{et al.}: Motion Comfort Optimization for Autonomous Vehicles}

\maketitle

\begin{abstract}
This article outlines the architecture of autonomous driving and related complementary frameworks from the perspective of human comfort. The technical elements for measuring Autonomous Vehicle (AV) user comfort and psychoanalysis are listed here. At the same time, this article introduces the technology related to the structure of automatic driving and the reaction time of automatic driving. We also discuss the technical details related to the automatic driving comfort system, the response time of the AV driver, the comfort level of the AV, motion sickness, and related optimization technologies. The function of the sensor is affected by various factors. Since the sensor of automatic driving mainly senses the environment around a vehicle, including ``the weather" which introduces the challenges and limitations of second-hand sensors in autonomous vehicles under different weather conditions. The comfort and safety of autonomous driving are also factors that affect the development of autonomous driving technologies. This article further analyzes the impact of autonomous driving on the user's physical and psychological states and how the comfort factors of autonomous vehicles affect the automotive market. Also, part of our focus is on the benefits and shortcomings of autonomous driving. The goal is to present an exhaustive overview of the most relevant technical matters to help researchers and application developers comprehend the different comfort factors and systems of autonomous driving. Finally, we provide detailed automated driving comfort use cases to illustrate the comfort-related issues of autonomous driving. Then, we provide implications and insights for the future of autonomous driving.

\end{abstract}
\begin{IEEEkeywords}
Autonomous driving, Autonomous vehicles, Sensors, Comfort, Driving safety, Psychoanalysis, Machine learning, Response time, Reaction time
\end{IEEEkeywords}

\section{Introduction}

\IEEEPARstart{C}urrently, the autonomous vehicles (AV) market is forecasted to reach 186.40 billion USD by 2030, and in 2021 alone, the AV market was worth 4 billion USD. Such high investments prove that AVs have become the primary revolution in connectivity technology and automation, and many operations have become reality \cite{newswire_2022}. Research on AVs is still developing rapidly since AVs  have emerged to provide travel services for humans. Recent research regarding AVs is focused on achieving significant improvement in the AV's safety, fuel consumption, efficiency, and comfort. Evaluation factors for AVs have also been examined and implemented in terms of safety and energy, but comfort seems to be the most difficult evaluation factor to define properly. This brings to attention a problem within the AV research community: unclear guidelines and standards on how to provide the best driving experience via a human-machine interaction system. The comfort of the customer for the AV involves not only the passenger but also the driver \cite{duan2022implementation}. The AVs are classified from the automation perspectives into five levels, 0-5. Level 0 refers to vehicles with no automation; level 5 represents fully autonomous vehicles, where drivers do not intervene. Level 2 indicates that drivers must control some main driving tasks while some can be partially autonomously. Vehicles have complete control to drive autonomously in levels 3 and 4, but drivers are allowed to intervene if needed. In levels 3 and 4, drivers can manage their time to relax for specific periods and drive at other times. As such, with each increasing level of automation, comfort becomes more essential \cite{bellem2016objective}.

\begin {figure} [htbp]
\centering
e\includegraphics[width=3.4in]{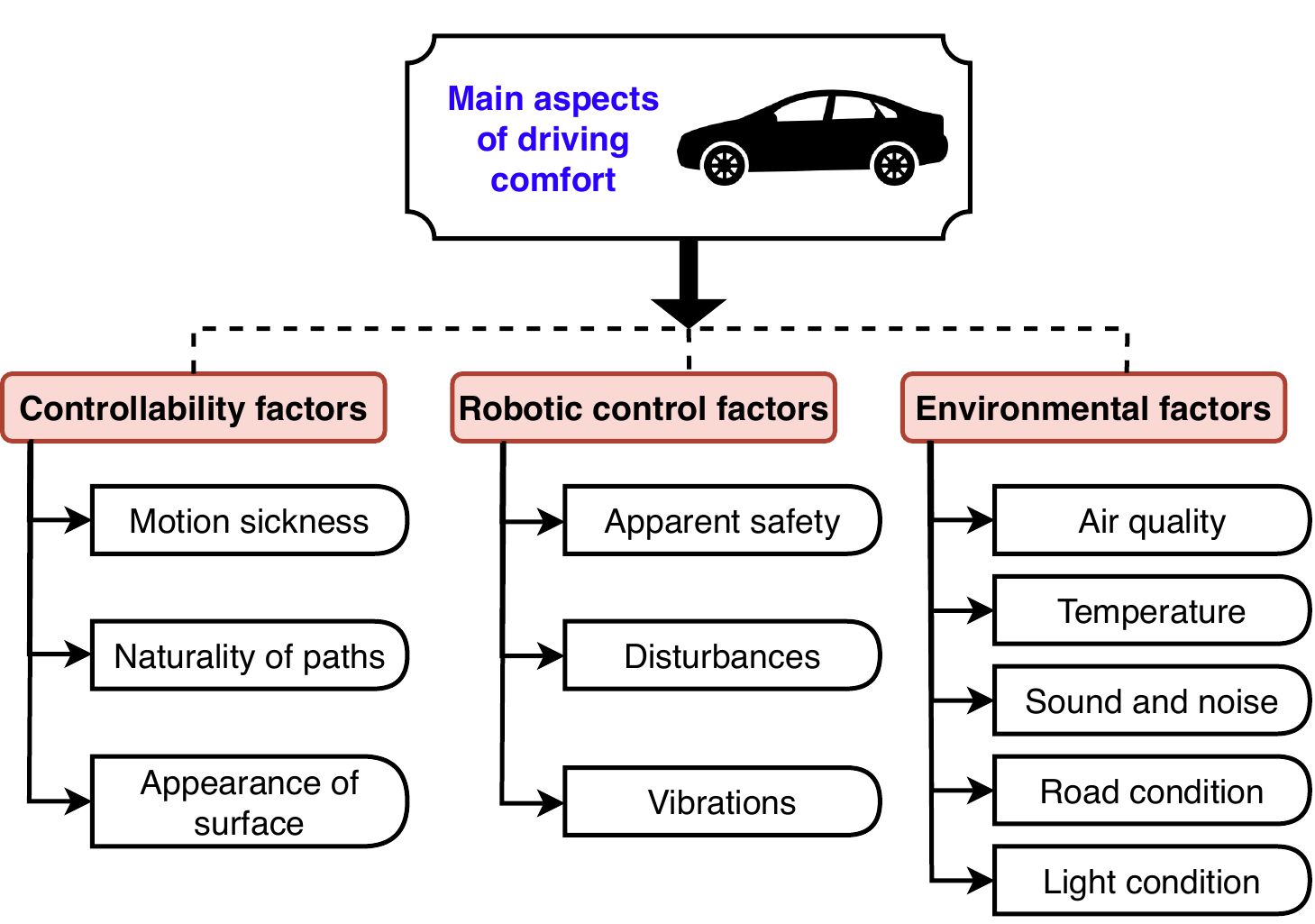}
\caption {Crucial concepts to consider for improving AV experience for consumers.}
\label{comfort_taxonomy}
\end {figure}

Comfort can be categorized by different factors as shown in Figure \ref{comfort_taxonomy}. Here, controllability factors represent the probabilistic attributes linking accidents to hazardous events based on the likelihood of the driver's ability to control the hazardous situation and thus evade harm. Robotic control factors represent system factors contributing to the movement of the car which involves the program and mechanical aspects that make it possible to control the car operations \cite{shah2019safe}, \cite{hakimi2018trust}, \cite{article}, \cite{articleww}. In general, the comfort level of AV users is challenging to eliminate or minimize because it is subjective. Different people will have different levels of comfort. Comfort of self-driving cars is impacted by many aspects, among which the reaction time of the self-driving car impacts the comfort of passengers both physiologically and psychologically \cite{arakawa2019psychophysical}. Note that the user's comfort level includes both the driver and the passengers of the self-driving car because once it is successfully implemented, the driver will now become a passenger. Unfortunately, when it comes to AV planning and design, as far as the performance of self-driving cars and user comfort of self-driving cars are concerned, user comfort is treated as an after-thought \cite{jing2020determinants}. The lack of extensive investigations of control strategies of AVs and their impact on the comfort factors of AV users and drivers contributed to such significant issues \cite{bae2019toward} \cite{zheng20223dop}, \cite{9699326} \cite{pal2022veriblock}. 

\begin {figure} [htbp]
\centering
\includegraphics [width=.9\columnwidth] {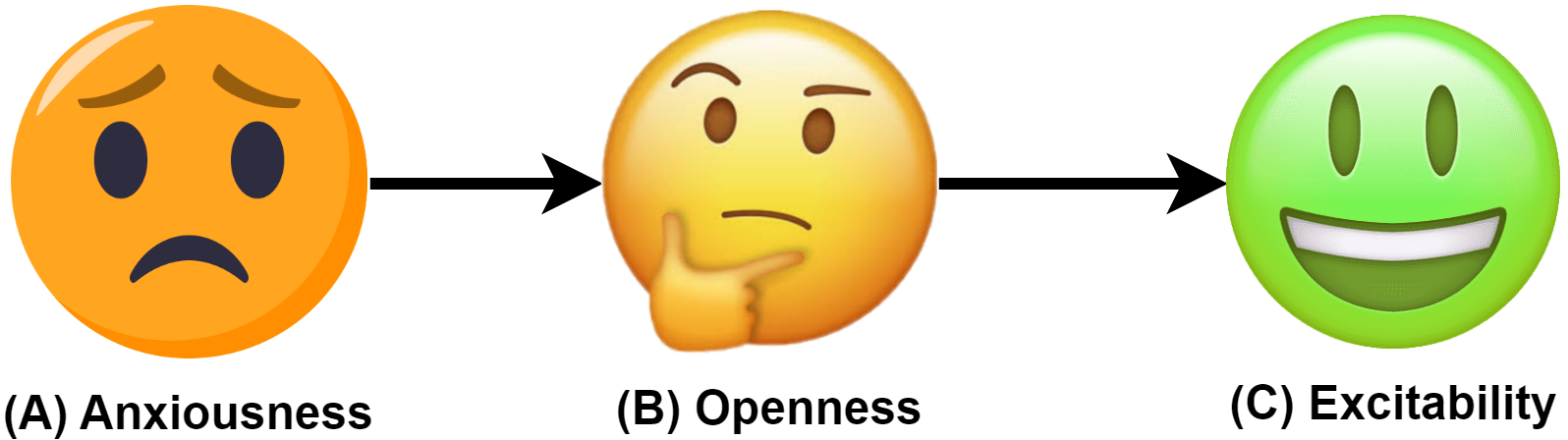}
\caption {A passenger's emotive state for using AVs based on the concept of HMI \cite{8699194}.}
\label{HMI}
\end {figure}

Further complicating matters regarding comfort for AVs is that there is no standard definition for it in the scientific community \cite{8699194}. Some researchers have categorized the comfort factors into either comfort or discomfort classes. As such, comfort would simply be defined as the absence of discomfort. Additionally, \cite{8699194} the authors discussed whether comfort factors should be treated as discomfort in the same significance or whether one of them is more important than the other. This problem is also relevant to the Human-Machine-Interaction (HMI) areas. Moreover, some literature divided the comfort factors into three levels of a human's expressive states, as seen by Figure \ref{HMI}. This is also applies towards riding/experiencing AVs for the first time \cite{avetisian2022anticipated}, \cite{antrobus2018trust}.  State A is a negative state, which relates to apprehension of riding AVs. These are the most complex and essential psychological and physiological factors for passengers to satisfy the required AV regulations since it is natural for people to get anxious when they are first using new devices. Usually, people have a significant level of anxiety if they do not have prior practice or use for the new technology or device \cite{8122757}, \cite{Sun2020ExploringPA}. This anxiety may accumulate in other forms of discomfort for the passenger \cite{7504447}, \cite{das2019youtube}. State B corresponds to the openness (willingness) of that passenger who wants to use the AV as an ordinary use case, which represents the main goal for autonomous vehicles. State C represents the positive state (excitability) when passengers get used to riding with AVs.

\begin{figure*}[!t]
\centering
\centerline{\includegraphics[height=11.5cm, width=18.5cm]{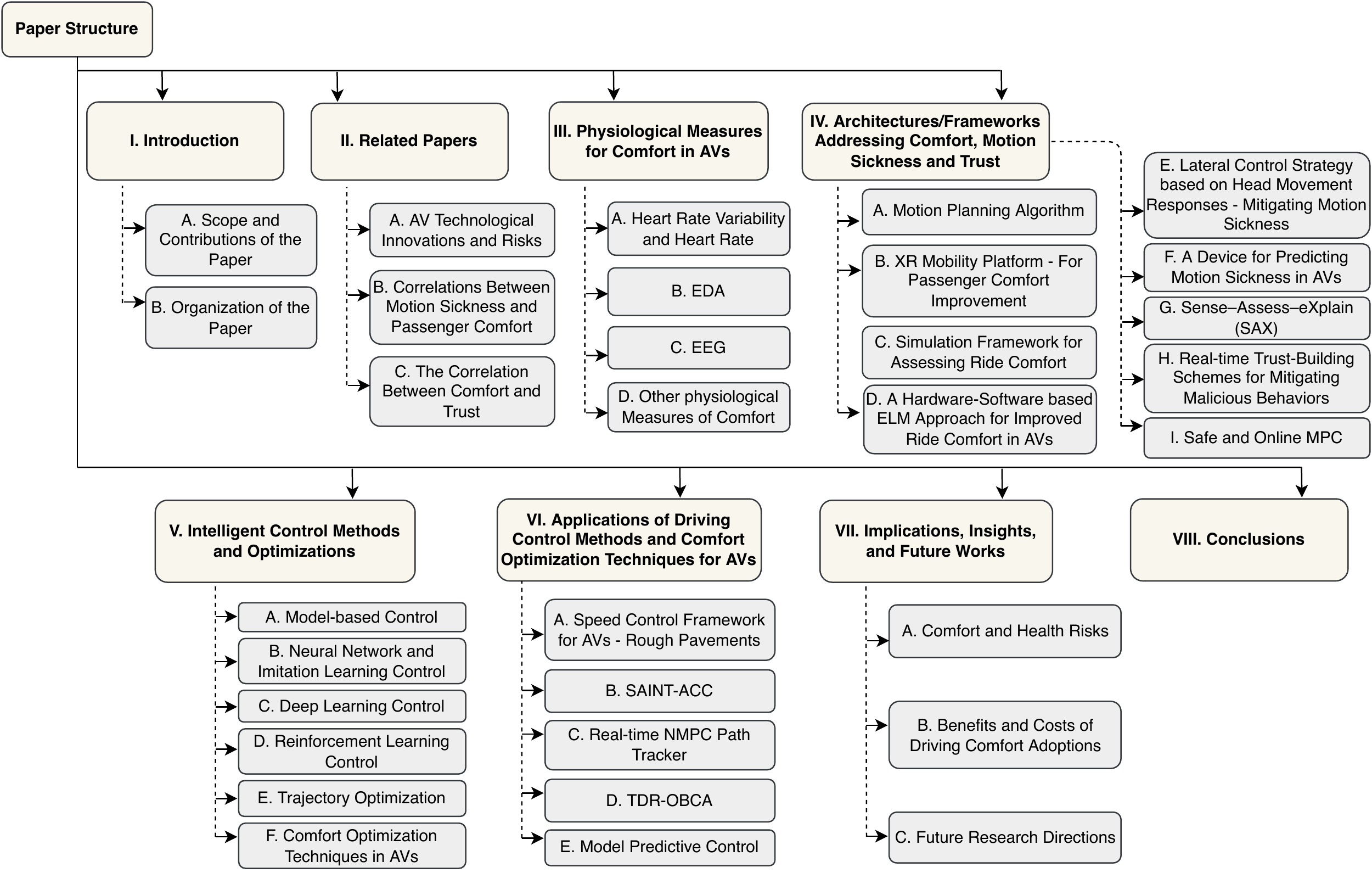}}
\caption{\textcolor{black}{A roadmap of the paper.}}
\label{fig:paperStructure}
\end{figure*}

When studying these topics, some researchers focused on a specific type of discomfort (e.g., motion sickness, anxiety) in relation to AVs. Out of all the types of discomforts relating to AVs, motion sickness is the most prevalent, as it has a huge impact on the consumer's riding experience, acceptance, and trust in the AV. What causes motion sickness has not yet been conclusively clarified, but according to a common theory, the cause is likely a sensory conflict; The AV's movement and its expectation (what we see) do not match. According to this theory, it should be able to help give the AV's occupants as precise information as possible about the impending movement – acceleration, curves, braking, and so on. Motion sickness can be identified in several ways, such as the presence of eye strain, nausea, headache, vomiting, sweating, and difficulty focusing. Such symptoms can be detected in AV drivers and passengers by checking and monitoring their physiological conditions. Motion sickness is one of the active AV research areas, with many strategies and frameworks being implemented to mitigate to achieve the AV's maximum comfort. 

\textcolor{black}{Safety and security are further crucial factors that affect the comfort of AVs. In terms of safety, AVs are designed to prevent accidents and minimize the risk of injury to passengers and road users. The safety features of AVs include advanced sensors, cameras, and machine learning algorithms that detect obstacles, pedestrians, and other vehicles in real-time. These features work together to ensure that the vehicle stays on the road and does not collide with other objects or vehicles. In terms of security, AVs are vulnerable to cyber attacks, which may compromise the safety and security of passengers in the vehicle. For instance, an adversary could take control of the vehicle and cause it to crash or divert it from its intended route. Thus, safety and security are critical factors that impact the comfort of AV users \cite{jing2020determinants}. Passengers must feel safe and secure while using these vehicles to enjoy a comfortable ride. Therefore, manufacturers must prioritize safety and security in the design and implementation of autonomous vehicles.
}

This article explores the various possibilities that affect AV user comfort based on the elements listed in technical aspects, biomedical sensors, and psychological analysis. It looks at problems that affect the driver's reaction time in an incident, the comfort level in AVs, the movement dysfunction, smoothness, and jerkiness, and whether autonomous driving is relaxing and safe. Moreover, the article also explores comfort-related factors of autonomous driving, such as the impact of safety plus self-driving car architectures. This article not only reviews the comfort of autonomous driving from a technical perspective but also involves the physiological conditions of autonomous driving. It also explores the impact and benefits and costs of autonomous driving from an economic perspective.


\subsection{Scope and Contributions of the Paper}

This work reviews and discusses topics of AVs with a focus on comfort and motion sickness \textcolor{black}{from the humans/users (i.e., drivers and passengers) perspective}. We cover a broad range of technical aspects of both topics in regards to some current frameworks that have been developed and some common questionnaires that have been proposed. The paper also covers movement dysfunction, smoothness, and jerkiness, as those factors impact both the comfort of AVs for the user as well as the level of motion sickness experienced. Our main contribution is an exploration of the comfort factors and motion sickness of AVs from a technical perspective. We examine current architectures and frameworks proposed for mitigating motion sickness for AVs. We also examine the correlation between trust and comfort for AVs, as trust is also a crucial topic of AVs that is still talked about. There has been discussion about trust in terms of safety and security, but little about whether trust and comfort are correlated. We provide arguments as to how and why trust and comfort for AVs are correlated. \textcolor{black}{Unlike existing studies that are summarized in Table \ref{tab:relatedworks}, the key contributions of this article are:
\begin{itemize}
    \item Providing a comprehensive and informative overview of the technical and non-technical elements that impact human comfort in autonomous driving.
    \item Exploring the different comfort factors and motion sickness from a technical perspective, examining prominent frameworks and architectures proposed for addressing motion sickness, and analyzing the correlation between trust and comfort for AVs.
    \item Presenting a clear understanding of the challenges and limitations in assessing and improving the comfort of AV users and drivers, which is a missing piece of the existing literature.
\end{itemize}}


\subsection{Organization of the Paper}

\textcolor{black}{The roadmap and organization of this article are shown in Figure \ref{fig:paperStructure}.} Specifically, the document structure includes a discussion of similar work in Section \ref{relatedConcepts}. These include similar research areas and different aspects of AVs such as comfort factors, motion sickness, and incorporating trust in the AVs comfort factors. Then, the article discusses some of the most common physiological measurements used for comfort in Section \ref{physio}. In Section \ref{archs-frameworks}, the article discusses some current frameworks and representative frameworks that address comfort, motion sickness, and trust in AVs. Then, Section \ref{Intelligent} discusses some control process models for enhancing comfort for AVs. Section \ref{Apps} also discusses some strategies, applications, and implementations done by some researchers to enhance comfort, trust, and other aspects of AVs. Furthermore, in Section \ref{insights} \textcolor{black}{security risks and} health issues, best practices in designing safe AVs, and future research directions are provided. Section \ref{insights} further identifies important and growing research opportunities for autonomous driving computing technology. Last, the conclusion is in Section \ref{conclusions}.

\section{Related Papers}\label{relatedConcepts}

\begin{table*}[!h]
\tiny
\centering
\caption{Summary of related works}
\label{tab:relatedworks}
\begin{tabular}{@{}lccccccl@{}}
\toprule
\multicolumn{1}{c}{\textbf{Refs. (Author)}} & \textbf{\begin{tabular}[c]{@{}c@{}}Year of \\ Publication\end{tabular}} & \textbf{\begin{tabular}[c]{@{}c@{}}Comfort\\ Background\end{tabular}} & \textbf{\begin{tabular}[c]{@{}c@{}}Architectures\\ and Frameworks\end{tabular}} & \textbf{\begin{tabular}[c]{@{}c@{}}Control\\ Methods\end{tabular}} & \textbf{\begin{tabular}[c]{@{}c@{}}Comfort \\ Optimization\end{tabular}} & \textbf{\begin{tabular}[c]{@{}c@{}}Research Implications\\ and Trends\end{tabular}} & \textbf{Overview}                                                                                                                                                                                                         \\ \hline \hline
Cunningham et al. \cite{cunningham2015autonomous}                          & 2015                                                                    & \checkmark{}                                     & \xmark{}                                                       & \xmark{}                                          & \xmark{}                                                & \checkmark{}                                                   & \begin{tabular}[c]{@{}l@{}}Documented some of the human factors challenges for transitioning \\ from manually driven to self-driving vehicles.\end{tabular}                                                                \\ \midrule
Elbanhawi et al. \cite{elbanhawi2015passenger}                           & 2015                                                                    & \checkmark{}                                     & \checkmark{}                                               & \xmark{}                                          & \xmark{}                                                & \xmark{}                                                           & \begin{tabular}[c]{@{}l@{}} Tested comfort criteria and discussed some open path planning questions \\ from a passenger comfort perspective.\end{tabular}                                                   \\ \midrule
Koopman et al. \cite{koopman2017autonomous}                             & 2017                                                                    & \xmark{}                                             & \checkmark{}                                               & \xmark{}                                          & \xmark{}                                                & \checkmark{}                                                   & \begin{tabular}[c]{@{}l@{}}Identified the challenges and safety risks associated with inspecting reliable AV systems and \\ how they relate to other safety areas\end{tabular}                                            \\ \midrule
Fleetwood et al. \cite{fleetwood2017public}                           & 2017                                                                    & \xmark{}                                             & \xmark{}                                                       & \xmark{}                                          & \xmark{}                                                & \checkmark{}                                                   & \begin{tabular}[c]{@{}l@{}}Reviewed critical public health implications of AV and analyzed the key important \\ ethical issues inherent in AVs design\end{tabular}                                                         \\ \midrule
Crayton et al. \cite{crayton2017autonomous}                             & 2017                                                                    & \xmark{}                                             & \xmark{}                                                       & \xmark{}                                          & \xmark{}                                                & \checkmark{}                                                   & \begin{tabular}[c]{@{}l@{}}Discussed relations between technological innovations in AV and public health risks. Also\\ talks about the relevant health implications of AV policies\end{tabular}                          \\ \midrule
Kelley et al. \cite{kelley2017public}                              & 2017                                                                    & \xmark{}                                             & \xmark{}                                                       & \xmark{}                                          & \checkmark{}                                        & \checkmark{}                                                   & \begin{tabular}[c]{@{}l@{}}Analyzed public health challenges in the USA and reviews critical factors for determining \\ AV benefits and risks to public safety\end{tabular}                                               \\ \midrule
Noy et al. \cite{noy2018automated}                                 & 2018                                                                    & \xmark{}                                             & \checkmark{}                                               & \checkmark{}                                  & \xmark{}                                                & \xmark{}                                                           & Discussed some AV safety concerns                                                                                                                                                                     \\ \midrule
Morando et al. \cite{morando2018studying}                             & 2018                                                                    & \xmark{}                                             & \checkmark{}                                               & \xmark{}                                          & \xmark{}                                                & \checkmark{}                                                   & Addressed possible safety risks and impacts of AVs as they continue to advance                                                                                                                                              \\ \midrule
Burkhard et al. \cite{burkhard2018requirements}                            & 2018                                                                    & \checkmark{}                                     & \xmark{}                                                       & \checkmark{}                                  & \checkmark{}                                        & \checkmark{}                                                   & \begin{tabular}[c]{@{}l@{}}Simulated an AV driving that aims to examine the occupant's movements in a vehicle. \\The goal was to address ISO-2621 standard and extend it.\end{tabular}                                 \\ \midrule
Telpaz et al. \cite{telpaz2018approach}                              & 2018                                                                    & \checkmark{}                                     & \checkmark{}                                               & \checkmark{}                                  & \xmark{}                                                & \checkmark{}                                                   & Introduced a new method to identify and classify passengers' discomforts                                                                                                                                                 \\ \midrule
Taeihagh et al. \cite{taeihagh2019governing}                            & 2019                                                                    & \xmark{}                                             & \xmark{}                                                       & \xmark{}                                          & \xmark{}                                                & \checkmark{}                                                   & \begin{tabular}[c]{@{}l@{}}Reviewed governance strategies formed in response to AV evolution and discussed\\ technological risks of AVs on different areas\end{tabular}                                                  \\ \midrule
Paddeu et al. \cite{paddeu2020passenger}                              & 2020                                                                    & \checkmark{}                                     & \checkmark{}                                               & \xmark{}                                          & \xmark{}                                                & \xmark{}                                                           & Investigated the potential correlation between comfort level and  passengers' confidence in the AV                                                                                                                                                    \\ \midrule
Yurtsever et al. \cite{yurtsever2020survey}                           & 2020                                                                    & \xmark{}                                             & \checkmark{}                                               & \xmark{}                                          & \xmark{}                                                & \xmark{}                                                           & \begin{tabular}[c]{@{}l@{}}Discussed challenges of autonomous driving and covers the technical factors of automated \\driving. Also covers some issues in high-level system architectures and methodologies\end{tabular} \\ \midrule
Rojas et al. \cite{rojas2020autonomous}                               & 2020                                                                    & \xmark{}                                             & \checkmark{}                                               & \xmark{}                                          & \xmark{}                                                & \xmark{}                                                           & \begin{tabular}[c]{@{}l@{}}Discussed how proper policies and regulatory frameworks need to be implemented to handle\\ AVs before introducing AVs to the market.\end{tabular}                                             \\ \midrule
Sohrabi et al. \cite{sohrabi2020impacts}                             & 2020                                                                    & \xmark{}                                             & \xmark{}                                                       & \xmark{}                                          & \xmark{}                                                & \checkmark{}                                                   & Proposed a framework to systematically identify possible AV health risks                                                                                                                                                  \\ \midrule
Dam et al. \cite{dam2021review}                                 & 2021                                                                    & \xmark{}                                             & \checkmark{}                                               & \checkmark{}                                  & \checkmark{}                                        & \checkmark{}                                                   & Conducted a literature review on the new AVs directions and techniques                                                                                                                             \\ \midrule
Guo et al. \cite{guo2021can}                                 & 2021                                                                    & \checkmark{}                                     & \checkmark{}                                               & \xmark{}                                          & \xmark{}                                                & \checkmark{}                                                   & \begin{tabular}[c]{@{}l@{}} Experimented the impact of the existing prompt systems of vehicle driving \\ on passenger comfort and motion sickness \end{tabular}                                  \\ \midrule
Reuten et al. \cite{reuten2021assessing}                              & 2021                                                                    & \checkmark{}                                     & \xmark{}                                                       & \xmark{}                                          & \xmark{}                                                & \checkmark{}                                                   & Discussed correlation between general unpleasantness and specific symptoms for motion sickness                                                                                                                            \\ \midrule
De Winkel et al. \cite{de2022relating}                           & 2022                                                                    & \checkmark{}                                     & \xmark{}                                                       & \checkmark{}                                  & \xmark{}                                                & \checkmark{}                                                   & Investigated the potential correlation between the discomfort  with motion sickness                                                                                                                                                     \\ \midrule
Delmas et al. \cite{delmas2022effects}                              & 2022                                                                    & \checkmark{}                                     & \checkmark{}                                               & \xmark{}                                          & \checkmark{}                                        & \checkmark{}                                                   & \begin{tabular}[c]{@{}l@{}} Investigated several AV comfort factors, including road type, weather conditions \\, and traffic level \end{tabular}                                                       \\ \midrule
Daofei Li et al. \cite{li2022mitigating}                           & 2022                                                                    & \xmark{}                                             & \checkmark{}                                               & \xmark{}                                          & \xmark{}                                                & \checkmark{}                                                   & \begin{tabular}[c]{@{}l@{}} Introduced and analyzed a new vibration cue system to alert AV \\ passengers of predicted future AV moving and action   \end{tabular}                                                                                                                                          \\ \hline
This paper                                  & 2022                                                                    & \checkmark{}                                     & \checkmark{}                                               & \checkmark{}                                  & \checkmark{}                                        & \checkmark{}                                                   & \begin{tabular}[c]{@{}l@{}} Discussed existing architectures and frameworks that aim to enhance AVs comfort \\ while also discussing several relevant use-cases\end{tabular}                                  \\ \bottomrule
\end{tabular}
\end{table*}

Numerous studies attempt to discuss and cover the safety and human comfort factors for AVs. 
A study by Noy et al. \cite{noy2018automated} reviews the promises for improved safety that are espoused by proponents of automated driving and discusses considerations in the safety of automated driving, while the work by Yurtsever et al. \cite{yurtsever2020survey} discusses challenges of autonomous driving and covers the technical factors of automated driving. They also discuss some issues in high-level system architectures and methodologies, including planning and human-machine interfaces.

\subsection{AV Technological Innovations and Risks}
Further work by Rojas et al. \cite{rojas2020autonomous} and Fleetwood \cite{fleetwood2017public} focus on the AV risks to public safety. Specifically, Rojas et al. \cite{rojas2020autonomous} discuss the implications and risks of autonomous driving on public health and reviews policies and regulatory frameworks related to public safety and transportation, whereas Fleetwood \cite{fleetwood2017public} reviews critical public health implications of AV and analyzed the key important ethical issues inherent in AVs design.

As the technological evolution in the automated driving industry, the work by Crayton et al. \cite{crayton2017autonomous} reviews relations between technological innovations in AV and public health risks and also discusses the relevant health implications of AV policies. Similarly, the work by Koopman and Wagner \cite{koopman2017autonomous} identify the challenges and safety risks associated with validating dependable AV systems and how they relate to other safety areas. Additionally, another work by Kelly \cite{kelley2017public} analyzes public health challenges in the USA and reviews critical factors for determining AV benefits and risks to public safety.

Further research attempts such as the work by \cite{sohrabi2020impacts} considered developing a framework to systematically identify possible AV health risks. This work also reviewed AV's impact on public health and safety. Furthermore, the work by Curto et al. \cite{curto2021effects} and Morando et al. \cite{morando2018studying} address the safety implications and impacts of AVs. Notably, Curto et al. \cite{curto2021effects} cover the effects and ramifications of connected and autonomous vehicles on public safety, comfort, and security, while Morando et al. \cite{morando2018studying} analyzes the driving safety of AVs using a simulator.

Taeihagh and Lim in \cite{taeihagh2019governing} review governance strategies adopted in response to AV evolution and discussed technological risks of AVs in different areas, including safety and liability. Meanwhile, Cunningham and Regan \cite{cunningham2015autonomous} identify and analyze challenges related to transitioning from manually driven cars to AV and outline requirements to address them.

\subsection{Correlations Between Motion Sickness and Passenger Comfort}
The work of Guo et al. incorporates driving condition prompts to alleviate motion sickness and improve passenger comfort \cite{guo2021can}. The authors utilized a driving simulator to assess the influence of a developed Driving Condition Prompt (DCP) system on passengers' motion sickness and comfort. The authors claimed their experiments could improve passenger comfort when utilizing the vehicle's DCP systems and alleviate motion sickness. This system claims to provide promising enhancement for AV comfort and delivers sufficient guidelines for future HMI implementations for smart cars.

In another work \cite{li2022mitigating}, Li and Chen tackle the challenge of motion sickness, with semi- and fully automated vehicles. They develop a vibration cue system that uses tactile stimulation to inform passengers of future AV movement. Furthermore, this work analyzed some AV simulators' results of motion planning that aim to optimize vibration cueing time and patterns. The proposed approach utilized a vibration cue cushion-based system by evaluating the data and reactions of 20 participants. The results of the examined system showed a significant increase in passengers' understanding of the cues. The proposed approach claimed the motion sickness of the participants improved significantly, with results of an average success rate of 89\% for generating motion anticipation.

The work of Reuten et al. \cite{reuten2021assessing} also focuses on motion sickness. Specifically, the authors investigate whether there is a possible correlation between general unpleasantness and specific symptomatology for motion sickness. The article highlights a research gap in the symptomatology of motion sickness and unpleasantness, mainly that the correlation between them is unclear. After all, motion sickness is shown via feelings of unpleasantness, which range from slight discomfort to absolute dreadfulness. The authors also argue that there have been previous studies that have reported positive correlations between motion sickness symptomatology and measures of unpleasantness. However, the correlation-based investigation could hide potential deviations. The most significant manifesting symptom of this issue is vomiting, which is said to offer relief. For that reason, it will not be fair to rate the bad feeling of individuals in a similar fashion to how close they are to vomiting. The proposed work investigated a linear relationship to consider the bad feeling of passengers as potential symptoms progress. The suggested work divided their experiments into two phases; 1) the development of unpleasantness and other signs during continuous motion stimulation; 2) the development of unpleasantness progression of motion sickness symptoms. The authors studied the sickness rates in two published experimentations that deployed 20 to 30-minute motion sickness stimuli through virtual and natural motion. Each test included numerous sessions on different days, and the rates were obtained at 2 to 5 minutes. The experiment results inferred a positive correlation between motion sickness symptoms and unpleasantness, but still, there is a period of relief at the onset of nausea. However, there is a need for further studies and analysis before the generalization of procedure outcomes that used only two different scales in addition to the listed anomaly.

The authors of \cite{de2022relating} investigated the impact of discomfort factors through monitoring the motion sickness and associated factors that were expressed in (\textit{Experiment 1}) and (\textit{Experiment 2}). The listed experiments utilized the MISC test. \textit{Experiment 1} employed 15 participants and analyzed the discomfort levels and associated MISC test results with each level, all participants were males aged between 20 and 38 years old. \textit{Experiment 2} has utilized 17 participants whose data were analyzed at 60 minutes or until reaching level 6 of MISC. After those periods, a 10-minute break is given, followed by 30 minutes for either movement measurements or reaching MISC 6 with a 0.3 Hz oscillation frequency. After that, participants conducted a second experiment in a completely dark environment. The MISC at 30-second intervals is used to rate the level of sickness of participants in each session. In \textit{Experiment 1}, the authors found that the discomfort level is increased using the MISC score. The experiments also found that participants with headaches, warmth, severe dizziness, stomach awareness, and sweating symptoms are more associated with a high uncomfortable level. The authors also found that discomfort levels associated with each level of the MISC increase monotonously in \textit{Experiment 2}. While both experiments were in-depth and the results were strong, it should be noted that for both experiments, the participants were mostly male. The sample sizes of those experiments could have included more females. It would have been interesting to see the differences between genders in terms of discomfort.

The authors of \cite{dam2021review} conducted a literature review on some comfort factors using the PRISMA framework. The above article studied 41 articles for the period of 2006 to 2021. The literature review employed several keywords to identify articles discussing motion sickness, causation, and mitigation methods in human and auto-pilot driving modes. Several databases were employed in this work, such as IEEE Xplore, EBSCOhost, Scopus, and ACM Digital Library. Moreover, the mentioned article used some keyword combinations, such as ``\textit{autonomous vehicles and motion sickness}'', ``\textit{motion sickness and mitigation and autonomous vehicles}'', ``\textit{motion sickness and situation awareness and autonomous vehicles}.''
The human factors articles represented the majority of the literature review findings. Also, the above article included vehicle dynamics and control algorithms research. The article found that most studies utilized virtual reality (VR) headsets to simulate and collect data on AVs and drivers' behaviors in non-driving tasks, which cannot help. The findings above utilized driving routes that include causes of motion sickness. Moreover, some studies developed heuristic models involving mathematical prototypes for motion sickness and AV performance that aim to improve the dynamics and offer comfortable rides. The literature review also concluded that real-world driving simulators and VR headsets have better reliability than driving simulators and VR headsets.

The authors in \cite{schartmuller2020sick} take an interesting approach to investigating motion sickness in AVs. The hypothesis is that olfaction (which relates to the sense of smell) may play a role in reducing motion sickness non-invasively. They focused on special scents, such as lavender and ginger. The test track investigated the impacts of those smells on drivers and passengers in chauffeured drives. The results showed that drivers could be at risk due to those smells in the pre-test and post-test.

\subsection{The Correlation Between Comfort and Trust}
The work of \cite{paddeu2020passenger} has explored the potential connection between comfort and trust in AVs. The article examined specific characteristics of rider experiences in the AV environment that impact trust and comfort. The study involved 55 participants that rode a shared AV under experimental conditions at a test zone. Each experiment employed 2 participants in four trips, accompanied by a researcher and safety operative. Two scenarios were presented for each trip, the direction (forward/backward) and maximum vehicle speed. Each participant rated the `comfort' and 'trust' factors after each trip. The investigation found a significant connection between trust and comfort factors statistically. Moreover, other studies explored AVs' trust, such as \cite{inproceedings}, \cite{horsting2022adapting}, and \cite{article9090}.

The work of \cite{delmas2022effects} investigated the comfort factor, considering other characteristics, such as weather conditions, road type, and traffic. That study assumed participants drove under level 3 of autonomy using various situations, including congested traffic, highways, heavy rain, and following defined speeds. The participants rated their perceived comfort level as they were driving an AV. The outcomes of the experimentation showed unfavorable changes in comfort in driving on downtown and highway under various conditions, such as heavy rain and traffic. At the same time, the experiment showed that decelerating the car's speed improved driving comfort. Later, the experiment studied four profiles 1) adversity to speed reduction, 2) trust in automation, 3) mistrust in automation, and 4) risk averse.

Comfort in AVs has also been studied using some more novel techniques, as in the case of\cite{su2021study}. The authors noted that human comfort in AVs is still not discussed often, and it is a crucial aspect of user acceptance of AVs. The argument is that existing studies relating to AV comfort only concentrated on physical factors such as sitting posture, vibration, and noise. Fewer studies investigated some psychological factors, for which this article aims to provide sufficient details and discussions. The authors attempt to define being comfortable as having no discomfort since it has a dominant effect on human comfort. The authors carried out their study using a driving simulator. The work investigated the issue in a simulated AV environment. Furthermore, a collection of AV trips that include motions have been experimented (27 videos total). The collected data included the physiological signals and comfort score. The listed work added a pressing button that enabled participants to express comfortable levels through the press strengths while riding the vehicle. Pushing the button referred to discomfort, while not pushing indicated the comfortable. What is unique about this work is that the authors used wearable sensors to collect physiological signals. The experiment also involved different types of roads (e.g city roads, highway roads, and mountain/rural roads) and different types of driving styles (gentle, aggressive, and normal). The listed videos lasted 3 to 5 minutes, and the participants were asked to express their feelings after watching each video. The experiment participants' feelings were measured and monitored by dividing the test into many sessions. Each session ran for 75 minutes per day for each participant. The authors also used a Support Vector Machine (SVM) for comfort level detection accuracy, and the SVM achieved a 71\% accuracy detection rate. The authors' experiment was highly detailed and the results were comprehensive, but the study suffers from sample size issues, a lack of diversity regarding participants, and generalizability regarding the machine method used; Similarly, the work of \cite{niermann2020measuring} measured sitting position, heartbeat, discomfort, and skin conductance through deploying sensors.

The work \cite{burkhard2018requirements} attempts to address comfort in AVs with a different approach. Unlike the work discussed in this section thus far, the authors actually address the ISO-2631 standard, which evaluates the impact of environmental shakes on comfort, health, and efficiency. The authors proposed and experimented with a modified ISO-2621 standard. 
This work utilized a body measurement system (BMS) for documenting occupant movements while driving in actual and simulated traffic. The used safety head cap is equipped with a sensor, while another is hooked to the seat rail. The proposed modified model evaluated the head movement during the experiments. The outcomes of this model found that measuring the occupant's head can improve the objectification of driving comfort for an AV.

The work \cite{telpaz2018approach}, \cite{cohen2020identifying} focused on the discomfort of AV passengers. The authors assessed the discomfort level by collecting measurements of passenger feelings while driving in real-time on the road. The main goal of the work is to identify the passenger discomfort levels using multimodal. The above work ran image processing techniques using recorded data of the outward-looking camera, location, and routing to analyze the comfort and discomfort. The discomfort is defined as adverse changes in passengers' emotions. Those negative changes can happen in (1) unwanted body moving, such as a sharp brake, or (2) perceptions of exterior risks, such as a car being too close to a leading car. The main idea of the listed work was to recognize potential driving scenarios that may cause discomfort and classify those scenarios in real-time. Ten sessions of rides with two participants were conducted in the experiment. Two passengers were involved in this experiment, one sitting in the front seat and the other sitting behind the driver. The experimentation was carried out in an urban area of San Francisco. Participants drove two pre-designed paths using two distinct driving manners (1) driving in the AV environment with caution and (2) driving in the non-AV environment without caution. Drivers rated their driving discomfort and comfort in real-time. The driving was in real-time measurement using a scale of 1 to 10. One refers to the lowest comfort level, and 10 indicates the highest level. Also, a logistic regression model was used to analyze the participant discomfort readings. The experimental results showed that the perception of risk is critical to a motorist's pressure and discomfort.


\section{Physiological Measures for Comfort in AVs}\label{physio}

Using various physiological measures for assessing comfort in AVs is beneficial because they allow for a closer and more objective analysis of the user for the AV. We can compare several physiological measures against other potential factors of AV driving. Numerous studies have undertaken this approach, with a similar consensus being that it may be easier to quantify discomfort, rather than comfort itself. It is not easy to assess the AV driver's comfort precisely because it's subjective. Moreover, measuring the AV drivers' and passengers' comfort and discomfort in real-time is not pleasant for them and might result in wrong readings and feelings. So, for measuring comfort for AV driving, we need non-intrusive and objective discomfort measurement systems that can be utilized for adapting to the AV's driving style and make sure the driver and passengers are relaxed \cite{9040568}, \cite{info11080390}. Thankfully, recent technological advancements such as wearable sensors are able to give non-intrusive physiological measurements. Some examples of these are in Table \ref{tab:SensorsPhysiological}. Note that the Empatica E4 and Microsoft Band 2 have been used in multiple studies pertaining to motion sickness and comfort in AVs.
\begin{table}[h]
\centering
\caption{A list of sensors and the physiological measurements they capture.}
\label{tab:SensorsPhysiological}
\begin{tabular}{@{}
>{\columncolor[HTML]{FFFFFF}}l 
>{\columncolor[HTML]{FFFFFF}}l 
>{\columncolor[HTML]{FFFFFF}}c 
>{\columncolor[HTML]{FFFFFF}}l @{}}
\toprule
{\color[HTML]{222222} \textbf{Device}} & {\color[HTML]{222222} \textbf{Type}} & {\color[HTML]{222222} \textbf{Release Year}} & {\color[HTML]{222222} \textbf{\begin{tabular}[c]{@{}l@{}}Physiological \\ Signals\end{tabular}}} \\ \midrule
{\color[HTML]{222222} Apple Watch 7} & {\color[HTML]{222222} Smartwatch} & {\color[HTML]{222222} 2021} & {\color[HTML]{222222} BVP, ECG, SpO2} \\ \hline
{\color[HTML]{222222} Fitbit Charge 5} & {\color[HTML]{222222} Smartband} & {\color[HTML]{222222} 2021} & {\color[HTML]{222222} EDA} \\ \hline
{\color[HTML]{222222} \begin{tabular}[c]{@{}l@{}}Samsung \\ Galaxy Watch 4\end{tabular}} & {\color[HTML]{222222} Smartwatch} & {\color[HTML]{222222} 2021} & {\color[HTML]{222222} BVP} \\ \hline
{\color[HTML]{222222} Huawei Watch 3} & {\color[HTML]{222222} Smartwatch} & {\color[HTML]{222222} 2021} & {\color[HTML]{222222} HR} \\ \hline
{\color[HTML]{222222} Fitbit Sense} & {\color[HTML]{222222} Smartwatch} & {\color[HTML]{222222} 2020} & {\color[HTML]{222222} ECG, EDA} \\ \hline
{\color[HTML]{222222} \begin{tabular}[c]{@{}l@{}}Samsung \\ Galaxy Watch 3\end{tabular}} & {\color[HTML]{222222} Smartwatch} & {\color[HTML]{222222} 2020} & {\color[HTML]{222222} BVP} \\ \hline
{\color[HTML]{222222} Apple Watch 5} & {\color[HTML]{222222} Smartwatch} & {\color[HTML]{222222} 2019} & {\color[HTML]{222222} ECG} \\ \hline
{\color[HTML]{222222} Fossil Gen 5} & {\color[HTML]{222222} Smartwatch} & {\color[HTML]{222222} 2019} & {\color[HTML]{222222} BVP} \\ \hline
{\color[HTML]{222222} \begin{tabular}[c]{@{}l@{}}Garmin \\ Fenix 6X Pro\end{tabular}} & {\color[HTML]{222222} Smartwatch} & {\color[HTML]{222222} 2019} & {\color[HTML]{222222} BVP, SpO2} \\ \hline
{\color[HTML]{222222} \begin{tabular}[c]{@{}l@{}}Samsung \\ Galaxy Watch\end{tabular}} & {\color[HTML]{222222} Smartwatch} & {\color[HTML]{222222} 2019} & {\color[HTML]{222222} BVP} \\ \hline
{\color[HTML]{222222} Polar OH1} & {\color[HTML]{222222} Armband} & {\color[HTML]{222222} 2019} & {\color[HTML]{222222} BVP} \\ \hline
{\color[HTML]{222222} \begin{tabular}[c]{@{}l@{}}Garmin \\ HRM-DUAL\end{tabular}} & {\color[HTML]{222222} Chest strap} & {\color[HTML]{222222} 2019} & {\color[HTML]{222222} ECG} \\ \hline
{\color[HTML]{222222} Muse 2} & {\color[HTML]{222222} Headband} & {\color[HTML]{222222} 2019} & {\color[HTML]{222222} BVP, EEG, SpO2} \\ \hline
{\color[HTML]{222222} Fitbit Charge 3} & {\color[HTML]{222222} Fitband} & {\color[HTML]{222222} 2018} & {\color[HTML]{222222} HR} \\ \hline
{\color[HTML]{222222} \begin{tabular}[c]{@{}l@{}}Garmin Vivo\\ Active 3 Music\end{tabular}} & {\color[HTML]{222222} Smartwatch} & {\color[HTML]{222222} 2018} & {\color[HTML]{222222} HR} \\ \hline
{\color[HTML]{222222} Oura Ring} & {\color[HTML]{222222} Smart ring} & {\color[HTML]{222222} 2018} & {\color[HTML]{222222} HR, HRV, SKT} \\ \hline
{\color[HTML]{222222} Moodmetric} & {\color[HTML]{222222} Smart ring} & {\color[HTML]{222222} 2017} & {\color[HTML]{222222} EDA} \\ \hline
{\color[HTML]{222222} DREEM} & {\color[HTML]{222222} Headband} & {\color[HTML]{222222} 2017} & {\color[HTML]{222222} BVP, EEG, SpO2} \\ \hline
{\color[HTML]{222222} Polar H10} & {\color[HTML]{222222} Chest strap} & {\color[HTML]{222222} 2017} & {\color[HTML]{222222} ECG} \\ \hline
{\color[HTML]{222222} VitalPatch} & {\color[HTML]{222222} Chest patch} & {\color[HTML]{222222} 2016} & {\color[HTML]{222222} ECG, SKT} \\ \hline
{\color[HTML]{222222} Emotiv Insight} & {\color[HTML]{222222} Headband} & {\color[HTML]{222222} 2015} & {\color[HTML]{222222} EEG} \\ \hline
{\color[HTML]{222222} Empatica E4} & {\color[HTML]{222222} Wristband} & {\color[HTML]{222222} 2015} & {\color[HTML]{222222} BVP, EDA, SKT} \\ \hline
{\color[HTML]{222222} Microsoft Band 2} & {\color[HTML]{222222} Smartband} & {\color[HTML]{222222} 2014} & {\color[HTML]{222222} BVP, EDA, SKT} \\ \bottomrule
\end{tabular}
\end{table}

The analyzed physiological signals in the study regarding the AV participants comfort, included the heart rate variability (HRV), systolic blood pressure (SBP), heart rate (HR), Electrodermal Activity (EDA), and EEG. Some studies used the Galvanic Skin Response (GSR) and Skin Conductance Level (SCL) as alternative terms for the EDA \cite{8502860}. Other physiological measurements such as eye-tracking have also been used as demonstrated in table \ref{tab:PhysioStudies}.

\begin{table}[h]
\centering
\caption{Some AV-related studies and physiological measurements used.}
\label{tab:PhysioStudies}
\scalebox{0.8}{
\begin{tabular}{@{}cccc@{}}
\toprule
\textbf{Ref. Author(s)} & \textbf{Year} & \textbf{\begin{tabular}[c]{@{}c@{}}Physiological \\ Raw Signals \\ Measured\end{tabular}} & \textbf{Finding(s)} \\ \midrule
Affani et al.\cite{s22051785} & 2022 & EEG & \begin{tabular}[c]{@{}c@{}} EEG headband - uses to \\ identify \\ a confidence \\ of AV drivers \end{tabular} \\ \hline
Niermann et al.\cite{niermann2020measuring} & 2021 & \begin{tabular}[c]{@{}c@{}}Eye-tracking\\ BVP\end{tabular} & \begin{tabular}[c]{@{}c@{}} Pupil diameter used to \\ predict discomfort \end{tabular} \\ \hline
Gruden et al.\cite{s21020550} & 2021 & EGG & \begin{tabular}[c]{@{}c@{}} Use EGG to identify \\ severity of AV \\ rider sickness \end{tabular} \\ \hline
Dillen et al. \cite{dillen2020keep} & 2020 & \begin{tabular}[c]{@{}c@{}}GSR\\ HR\\ HRV\\ Eye-tracking\end{tabular} & \begin{tabular}[c]{@{}c@{}} GSR and HR - identify \\ and measure the magnitude \\ of the AV acceleration and jerk \\ \\ GSR predicted comfort and anxiety\\ \\ No significant correlations \\ between trust and skin conductance \\ regarding passenger trust for the AV\end{tabular} \\ \hline
\begin{tabular}[c]{@{}c@{}}Radhakrishnan et al.\cite{info11080390} \end{tabular} & 2020 & \begin{tabular}[c]{@{}c@{}}HRV\\ EDA\end{tabular} & \begin{tabular}[c]{@{}c@{}} SCR is more accurate than \\ HRV-based reading \\ to scenarios that \\ induce discomfort\end{tabular} \\ \hline
Beggiato et al. \cite{article4556} & 2019 & \begin{tabular}[c]{@{}c@{}}HR\\ EDA\end{tabular} & \begin{tabular}[c]{@{}c@{}}Decreased HR during \\ uncomfortable situations\\ \\ No notable effects for EDA\\ \\ Lowering eye blinks in\\ discomfort circumstances, \\ but pupil diameter increased\end{tabular} \\ \hline
Beggiato et al. \cite{articletttt} & 2018 & \begin{tabular}[c]{@{}c@{}}HR\\ HRV\\ SCL\\ Eye-tracking\end{tabular} & \begin{tabular}[c]{@{}c@{}}Reduced eye-blink rate \\ during discomfort\\ \\ HR- decreased significantly during \\ discomfort periods\end{tabular} \\ \hline
Shahrdar et al. \cite{8448945} & 2018 & EEG & \begin{tabular}[c]{@{}c@{}} Classified AVs EEG readings \\ using neural networks, \\ SVM, and MLP \end{tabular} \\ \hline
Zhang et al. \cite{zhang2017webcam} & 2017 & HR & \begin{tabular}[c]{@{}c@{}}Proposed approach \\ can conveniently\\ measure HR with \\ acceptable accuracy\end{tabular} \\ \bottomrule
\end{tabular}}
\end{table}

\subsection{Heart Rate Variability and Heart Rate}
Heart Rate Variability (HRV) and Heart Rate (HR) and are two types of measures that have been used in studies involving AVs, driving simulations, and on-the-road driving studies. With these, both HR and HRV have been used to indicate levels of anxiety, cognitive measure, workload, and duty demands \cite{arakawa2021review}. However, there is a distinction between HR and HRV; HR is measured in BPM (beats per minute) and usually does not require exact times. HR focuses on average BPM while HRV focuses on measuring specific changes in time between successive heartbeats. The time duration between these heartbeats is measured in milliseconds and is referred to as the R-R interval or inter-beat interval (IBI).

Regarding their involvement in AV-related studies, some findings such as those from \cite{article4556} and \cite{articletttt}, indicated that HR decreased consistently during uncomfortable driving situations. Similarly, HRV was shown to decrease. Interestingly, in \cite{dillen2020keep}, the authors actually found that GSR and HR were significantly impacted by driving parameters such as acceleration and jerk, especially in the presence of another vehicle. These aspects lead to the HR increasing between 3.6 and 6.5 BPM. In another work by \cite{zhang2017webcam}, The work developed a real-time webcam-based detection system to measure drivers' HR and monitor their physical activities. The color variations of the skin's blood circulation were used to identify the HR. 

\subsection{EDA} 
The autonomic nervous system (ANS) signals are used to track the electric conductance changes in the skin, which represent the EDA. The sweat gland activity can be increased using the arousal of the sympathetic ANS activity that leads to high skin conductivity. For that reason, EDA can be utilized to measure the physiological arousal readings of participants in response to the external stimulus \cite{articletttt}.

EDA is utilized to monitor and read information for various AV-related studies, similar to HR and HRV. EDA information includes the workload, task difficulty, mental effort, skin conductance level (SCL) with higher arousal, alertness, emotional load, and stress \cite{dawson2017electrodermal}. However, EDA is not a reliable metric due to its sensitivity to various stimuli. The increase in SCL readings would be expected due to higher alertness and arousal prediction indicating discomfort. There have been some notable findings with this physiological measure as demonstrated by \cite{info11080390}, where the authors found that EDA values were more sensitive than HRV values in terms of scenarios that induced discomfort in AVs. Other examples come from \cite{inproceedings}, \cite{article9090}, \cite{8502860}, \cite{article88888}, and \cite{article00000}, which examined the correlation between EDA and trust, where a high level of EDA refers to a low level of trust. However, these claims were disproven in \cite{dillen2020keep}, where the authors indicated there were no significant correlations between trust and EDA. A similar case also occurred in Julius Hörsting et al. \cite{horsting2022adapting}, where the authors compared EDA among three groups in regard to trust and EDA for AVs. The goal was to investigate the correlation between AV speeds and participants' trust. The EDA was used as the physiological measurement. The study didn't find a significant correlation between the three groups using the EDA readings.

\subsection{EEG}
An electroencephalogram (EEG) allows us to measure the electrical activities in human brain. Applying electrodes to the individual's scalp is the way to get EEG readings. EEGs have been used in AV studies for a few aspects, mainly emotional state and stress; One example comes from \cite{8448945}, where the authors incorporate EEG as an indicator of the emotional state of a passenger in an AV simulator. Another example comes from \cite{9689603}, where the authors hypothesize that abnormalities in braking timing for AVs were reflected in the physiological signals. Upon experimenting and evaluation, the authors' results indicated that it was in fact possible to estimate the driver's abnormal braking by using the physiological signals before and after the braking timing.  In \cite{s22051785}, the authors developed a wearable EEG headband to measure stress-related brain activity during driving. Upon testing the proposed approach on 10 volunteers, results showed that manual driving had the highest stress on drivers. 

Perhaps the most notable work that uses EEG comes from \cite{minea2021advanced}, where the authors focus on designing a complex sensor for AVs, by equipping a semi-AV with an intricate sensor structure that can provide centralized data regarding EEGs of both passenger and driver of the AV. This would in turn transform the AV into a mobile sensor connected via the Internet. The authors design the proposed sensor such that the acquisition devices of EEG signals are installed in the head and backrests on car seats. Not only does this allow for maximum comfort, but the approach can be used as a way to help the transition between manually driven vehicles and AVs by bringing forth a system that monitors the well-being of the driver's location in AV and non-AV environments.

\subsection{Other Physiological Measures of Comfort}
Some other physiological measures of comfort for AVS involve eye-tracking. Studies by Drewitz et al. \cite{drewitz2020towards}, Dillen et al. \cite{dillen2020keep}, and Beggiato et al. \cite{article4556} \cite{articletttt}, all incorporated eye-tracking when assessing for comfort/discomfort. The work found that the diameter of the pupil is primarily on ambient light and depends on mental states. Niermann also found that the pupil diameter is the best factor for predicting and evaluating AV drivers' discomfort \cite{niermann2020measuring}. Additionally, Beggiato et al. found that there is a reduced eye-blinking rate during discomforting AV scenarios. Furthermore, \cite{beggiatomulti} explores the possibility of incorporating face-tracking strategies to monitor comfort for driver state monitoring regarding discomfort in AV driving. Another technique, \cite{s21020550}, was used to identify the potential correlations between increased amplitude recorded EGGs and the severity of self-reported sickness. 

\section{Architectures/Frameworks Addressing Comfort, Motion Sickness and Trust}\label{archs-frameworks}

\subsection{Motion Planning Algorithm}
Many researchers have proposed various frameworks that address comfort and motion sickness for AVs, with one example coming from \cite{9502537}, which proposes a novel concept to remediate car-sickness problems in motion planning, instead of motion control. This work uses a frequency shaping method to develop a motion planning algorithm that is commonly used for checking automobile sickness. The outcomes of the simulation and experiment showed a promising reduction by 21\% and 37\% of motion sickness dose value (MSDV). The ISO 2631-1:1997 standard referred to the used MSDV-based motion planning algorithm. The MSDV is defined as follows:

\begin{equation*} \begin{array}{c} {MSDV\ = \ \sqrt {\mathop \smallint \nolimits_0^T {{\left[ {\tilde{a}\left(t \right)} \right]}^2}dt} } \end{array} \tag{1} \end{equation*}

$T$ refers to the total exposure time of the acceleration and the stimulus acceleration weighted by function, as defined in ISO 2631-1:1997. 

\begin{figure}[h]
    \centering
	\includegraphics[width=3.5in]{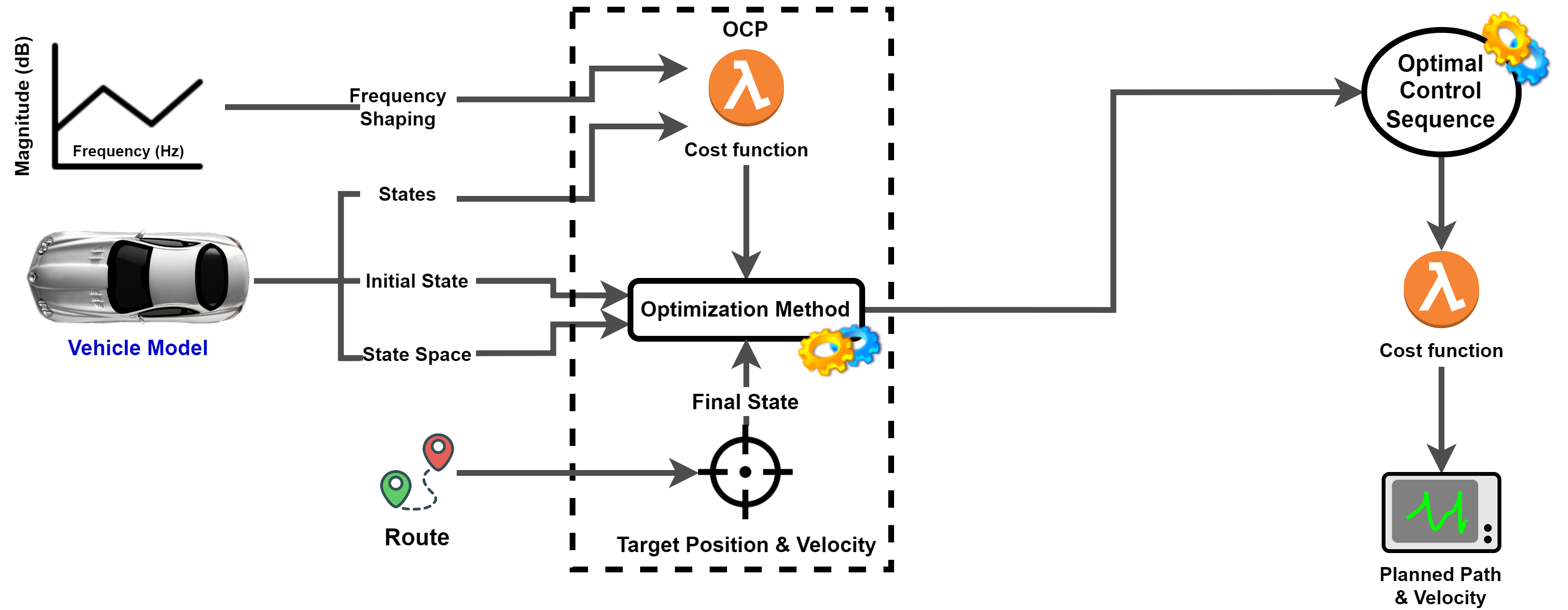}
	\caption {Illustration of the motion planning architecture \cite{9502537}.}
	\label{Arch1}
\end{figure}

Figure \ref{Arch1} \cite{9502537} shows the formulated motion planning algorithm similar to the optimal control problem (OCP). The planned vehicle trajectory is determined by minimizing the cost obtained by calculating the route planner's target positions and velocity profiles.

The actual mitigation happens in the motion planning phase, so there is no impact for the motion control to cause motion sickness. The need for fewer car parameters is a strength of the proposed approach, making it a sufficient and preferable algorithm for motion planning.

The proposed algorithm is designed to work in real-time, which requires initial and final conditions. The experiment set the initial speed vehicle to 1.7m/s for simulated congested urban driving conditions and 30 km/h for the main road. The experimentation assumed the vehicle was equipped with the optimal control algorithm to keep the car at the planned velocity on the intended path. The experiment included comparing the proposed framework to a polynomial-curve-based planning methodology, a benchmark algorithm. A polynomial-curved-based method is widely used for AV-based research. The polynomial function and time are used to calculate the AV path and velocity. The polynomial function parameter is defined in two steps and evaluated using the Root Means Squared Error (RMSE). The authors claimed that the experimental results showed that the proposed framework performed better performance than the benchmark algorithm in identifying comfort. They also stated that their method increased sickness alleviation by up to 37\%.

\textcolor{black}{Overall, motion planning algorithms can improve comfort by providing a smooth, predictable, and customizable ride experience based upon the following features.
\begin{itemize}
    \item Optimizing the trajectory to ensure that the ride is as smooth as possible and minimizing sudden accelerations, decelerations, and turns. This can alleviate the discomfort and motion sickness.
    \item Anticipating the behavior of other road users and adjusting the trajectory accordingly. This helps avoid sudden stops and maneuvers, reducing the likelihood of discomfort.
    \item Providing customizable driving preferences to take into account the driving preferences of individual AV users. For instance, if a user prefers a more conservative driving style, the AV can be programmed to adjust its trajectory accordingly.
    \item Providing an adaptive ride comfort by adjusting to changes in the environment, such as road conditions and traffic patterns. By adjusting the trajectory in real-time, the motion planning algorithms can ensure that the ride is as comfortable as possible, even in challenging conditions.
\end{itemize}}

\subsection{XR Mobility Platform - For Passenger Comfort Improvement}

\begin{figure}[h]
    \centering
	\includegraphics[width=0.45\textwidth]{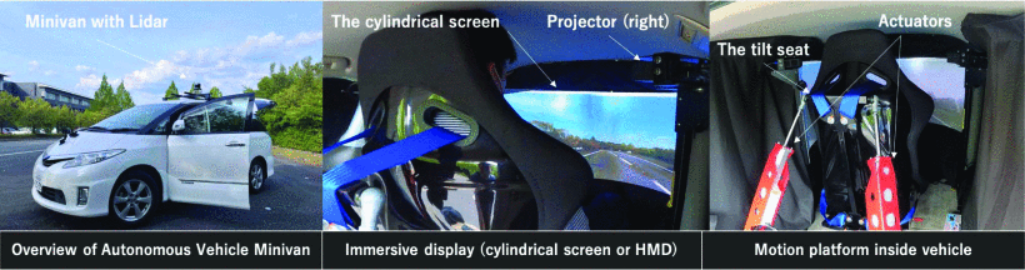}
	\caption {XR Mobility Platform Infrastructure \cite{inproceedings444}.}
	\label{XRArch}
\end{figure}

A multi-modal system introduced in the \cite{inproceedings444} could be mounted on the AV to improve passenger comfort. The work claimed that the drivers were freed from driving and became passengers, and the windshield and windows became dashboards to provide driving information. The goal was to implement technology for improving the passenger's comfort during auto-driving. A concept covered in this framework is comfort intelligence (CI), which focuses on passenger comfort in AVs and on both the positive and negative states of the passenger's feelings inside the AV. The authors claimed their approach is better than other approaches, particularly because it uses immersive displays with a controllable seat capable of tilting in all directions in the actual AV environment. The proposed method architecture is shown in Figure \ref{XRArch}.

The proposed framework includes several elements, as shown in Figure \ref{XRArch}. The Head Mounted Display (HMD) incorporates a high presence of VR inside the AV, and the dashboard camera records images. The passenger seat of the minivan's backspace equipped with the motion platform is used to monitor the motion. Numerous actuators are connected to the back of that seat to control tilting through a computer. The proposed framework also uses an open-source software called Autoware.AI, which deals with driving control. Autoware.AI gets behavioral data of the AV from sensors through a controller area network (CAN) bus. 

Furthermore, the proposed framework also uses two control methods to improve passenger comfort. The first one focuses on reducing car sickness. This strategy is supposed to reduce sensory conflict among the passenger by controlling the passenger's body movement directly. So, both the cylindrical screen and motion platform seat are tasked with reducing acceleration stimuli. With this approach, the passenger is supposed to feel fewer acceleration stimuli and what they do feel is similar to the driver's movement. The second control strategy focuses on reducing passenger movement. This second approach utilizes the HMD to control the visual movement and body of the passenger. The HMD used is the immersive HMD that is attached to the passenger's head. This helps creates an illusion that the passenger does not feel they are tilting, which helps the passenger feel less movement overall.

\textcolor{black}{Overall, the XR Mobility Platform has the potential to enhance passenger comfort by providing a more immersive, personalized, and interactive experience for passengers based on the following aspects.
\begin{itemize}
    \item VR simulations for different transportation scenarios, allowing passengers to experience their journey before it even begins. This can help to reduce anxiety and uncertainty, which may cause discomfort for some passengers.
    \item Passengers can create personalized virtual environments that suit their individual needs. For instance, they can adjust the lighting, temperature, and sound to enable a more comfortable environment.
    \item Real-time information for passengers about their journey, such as the location of their vehicle and any delays or disruptions. This helps reducing stress and improve the overall passenger experience.
    \item Interactive entertainment options for passengers, such as virtual games and movies, which may help to pass the time and distract them from any discomfort they may be experiencing.
\end{itemize}}

\subsection{Simulation Framework for Assessing Ride Comfort}
A simulation framework to evaluate comfort using collected data from vehicle dynamics has been proposed in \cite{12787}. The proposed simulation framework also generates a process for producing optimal comfort estimates by incorporating a Monte Carlo simulator and road surface model. The authors then develop a case study with the proposed simulation framework. The case study consists of a few real road sites that demonstrate the effectiveness of the proposed approach with actual data, and is able to achieve highly optimal comfort results. 

\begin{figure}[h] 
    \centering
 	\includegraphics[width=3.5in]{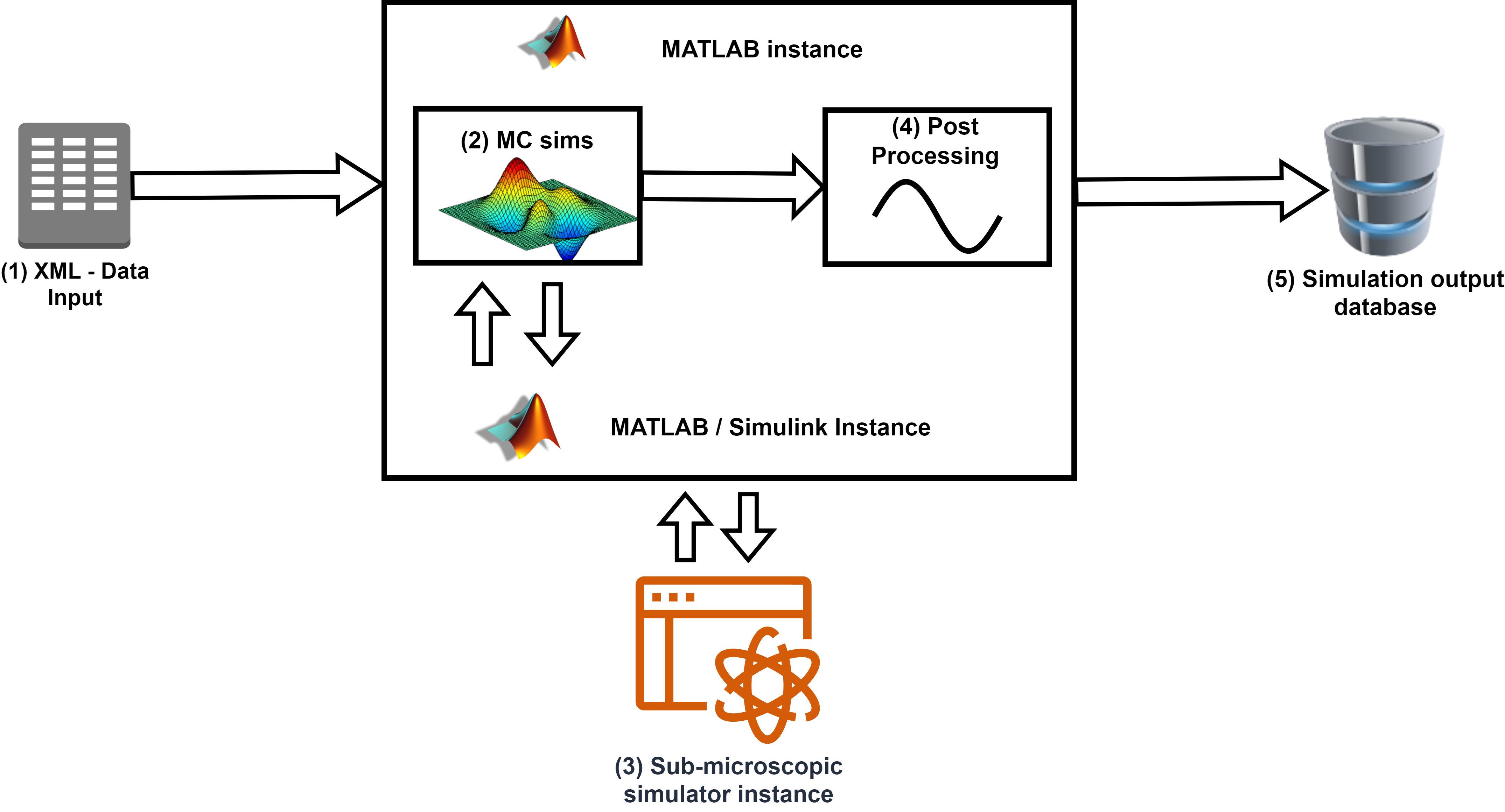}
	\caption{The proposed simulation framework generates a process for producing optimal comfort estimates by incorporating a Monte Carlo simulator and road surface model \cite{12787}.}
	\label{MonteCarlo}
\end{figure}

The work utilized specific frequency ranges that follow the ISO-2631 application domains to provide accurate comfort readings. As such, low frequency is between 0.5Hz and 80Hz in terms of health, comfort, and perception. Motion sickness varies in range from 0.1Hz to 0.5Hz; The proposed technique utilized a frequency weighting approach based on the ISO-2631 standard to identify frequency ranges and specific applications. The frequency weights have been selected based on the passenger position and the application type (i.e., sitting, standing, or recumbent). The overall conceptual architecture of the proposed method is shown in Figure \ref{MonteCarlo}.

The simulation framework is shown in Figure \ref{MonteCarlo}. An XML file is used to save the framework's configuration parameters, which provides flexibility to specify various test scenarios. The input parameters of the corresponding probabilistic are shown in block 1, and the automation functionality and the communication interface to the simulation environment are presented in block 2. The framework has been implemented using MATLAB/Simulink setting. MATLAB Simulink and CarMaker are both used as simulation software for set-up, input configuration, and data exportation (block 3). Block 4 included the dataset that utilizes to quantify comfort. Several methods are applied, such as bi-linear transformation, signal filtering, and final calculation. Finally, block 5 shows the storage of the comfort prediction. The authors chose to go with a generic framework so that input or output parameters can be easily replaced by different technologies. Moreover, other simulators compatible with MATLAB/Simulink can be used for the framework simulation instead of CarMaker. These changes will not negatively impact the performance of the proposed approach. The authors reported that their proposed simulation framework  found a relationship between the acceleration signal peaks and comfort.

\subsection{A Hardware-Software based ELM Approach for Improved Ride Comfort in AVs}
The authors in \cite{8852435} also focus on ride comfort. More specifically, the authors propose a hybrid hardware/software approach to improve AV ride quality. In addition, a driving style identification system is also integrated, as driving style is also taken into account in the proposed work. The proposed approach is utilized on a one-chip implementation to read driving comfort levels in real-time. So, the proposed framework categorizes the comfort-driving data into several types. The experiment used hierarchical cluster analysis (HCA) (unsupervised learning approach) and supervised extreme machine learning (ELM) to improve performance. HCA uses a bottom-up approach to measure the distance between two clusters, then merges the following two clusters at the next step. The Euclidean distance is used to evaluate HCA performance, similar to the nearest neighbor algorithm. The proposed framework uses Xilinx's Zynq-7000 programmable system-on-chip for real-time data processing and determining comfort levels. It also operated high operating frequencies and low latency times. The goal of this proposed approach was to alert drivers whenever passenger comfort was compromised.

\begin{figure*}[h]
    \centering
	\includegraphics[width=6in]{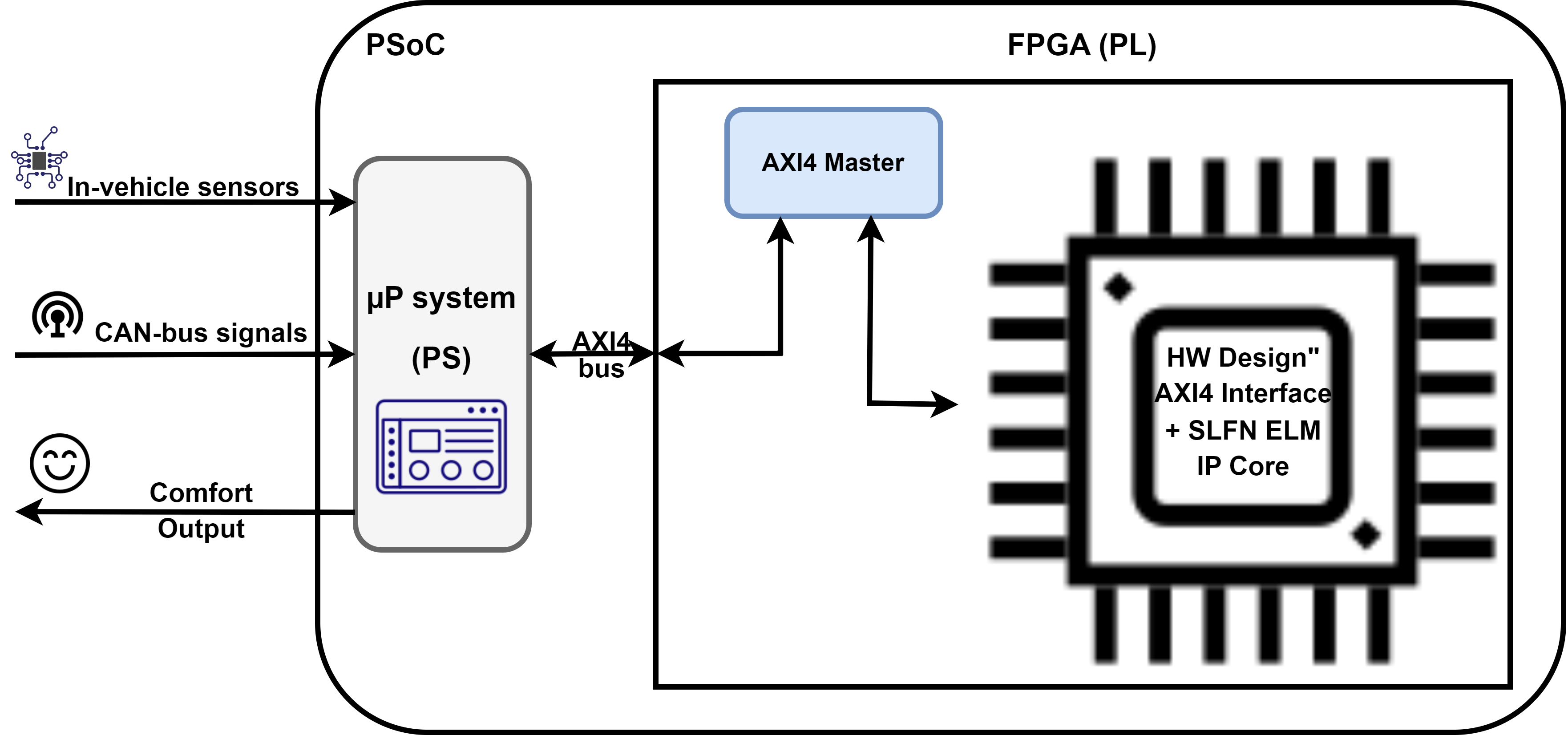}
	\caption {Hardware-Software based ELM approach for Improved Ride Comfort in AVs \cite{8852435}.}
	\label{hard-soft-arch}
\end{figure*}

This framework is featured by detecting non-continued circumstances, such as sudden lane shifts and moves that cause motion sickness and discomfort, especially if they are maintained over an extended time. The proposed work identifies the discomfort by recording the highest acceleration and jerk speeds, usually when aggressive lane changes occur. The discomfort can happen when values of acceleration and jerking are high, even for a short period. Figure \ref{hard-soft-arch} shows the framework components that handle the data exchange between the AV's CAN bus and the entire system. The driving style can be determined using the ELM-based model. To enable the system to work correctly and precisely must carefully follow specific configurations to integrate hardware with software. The proposed framework also utilized a dual-core ARM microprocessor with peripherals, so the software can run on a full-size FPGA where the ELM is deployed.
 
The experiment of the proposed approach showed a 95\% success rate of comfort after classifying driving data. The authors claimed that the proposed framework is helpful, especially the lack of driver advising-based comfort-improving solutions. They also indicated it could easily integrate the proposed approach with current auto line productions with some minor modifications to increase the comfort level of AV driving.

\subsection{Lateral Control Strategy based on Head Movement Responses - Mitigating Motion Sickness}
A new control strategy has been proposed to mitigate AV motion sickness, which uses head roll angle as a control variable to reduce lateral acceleration \cite{article2121}. The authors first develop a base model for driver and passenger head roll. The models of a driver and passenger are viewed as control subjects. This strategy employed a fuzzy logic control and head roll angle response to induce a corrective wheel angle. The proposed approach was run through via simulations, which showed a 3\% to 10\% reduction in motion sickness.

\textcolor{black}{Overall, lateral control strategies based on head movement responses are an effective way to mitigate motion sickness in autonomous vehicles by reducing the sensory conflict that causes it. By aligning the visual field with the physical motion of the vehicle, passengers can enjoy a more comfortable and enjoyable ride experience. For instance, if the vehicle turns left, the display can be shifted to the right, creating a compensatory visual stimulus that aligns with the physical motion of the vehicle. This mitigates the sensory conflict between the passengers' visual and vestibular systems, which is the primary cause of motion sickness. It is crucial to note, however, that this strategy may not work for all passengers and that individual differences in susceptibility to motion sickness should be taken into account.}

\subsection{A Device for Predicting Motion Sickness in AVs}
In \cite{Salter2019MotionSP}, the proposed framework is actually a device that could be used to predict user motion-sickness in real-time using motion, head tilt, and ambient conditions. The proposed approach heavily focuses on Non-Driving-Related-Tasks (NDRTs). Requirements for the device were the following:
\begin{itemize}
  \item Objective motion dose score
  \item A subjective sickness score
  \item Ambient temperature sensitive
  \item A position of the occupant head's
  \item Variability of the sensitivity tuning
  \item Real-time display and calculation
\end{itemize}

\begin{figure}[h]
    \centering
	\includegraphics[width=0.45\textwidth]{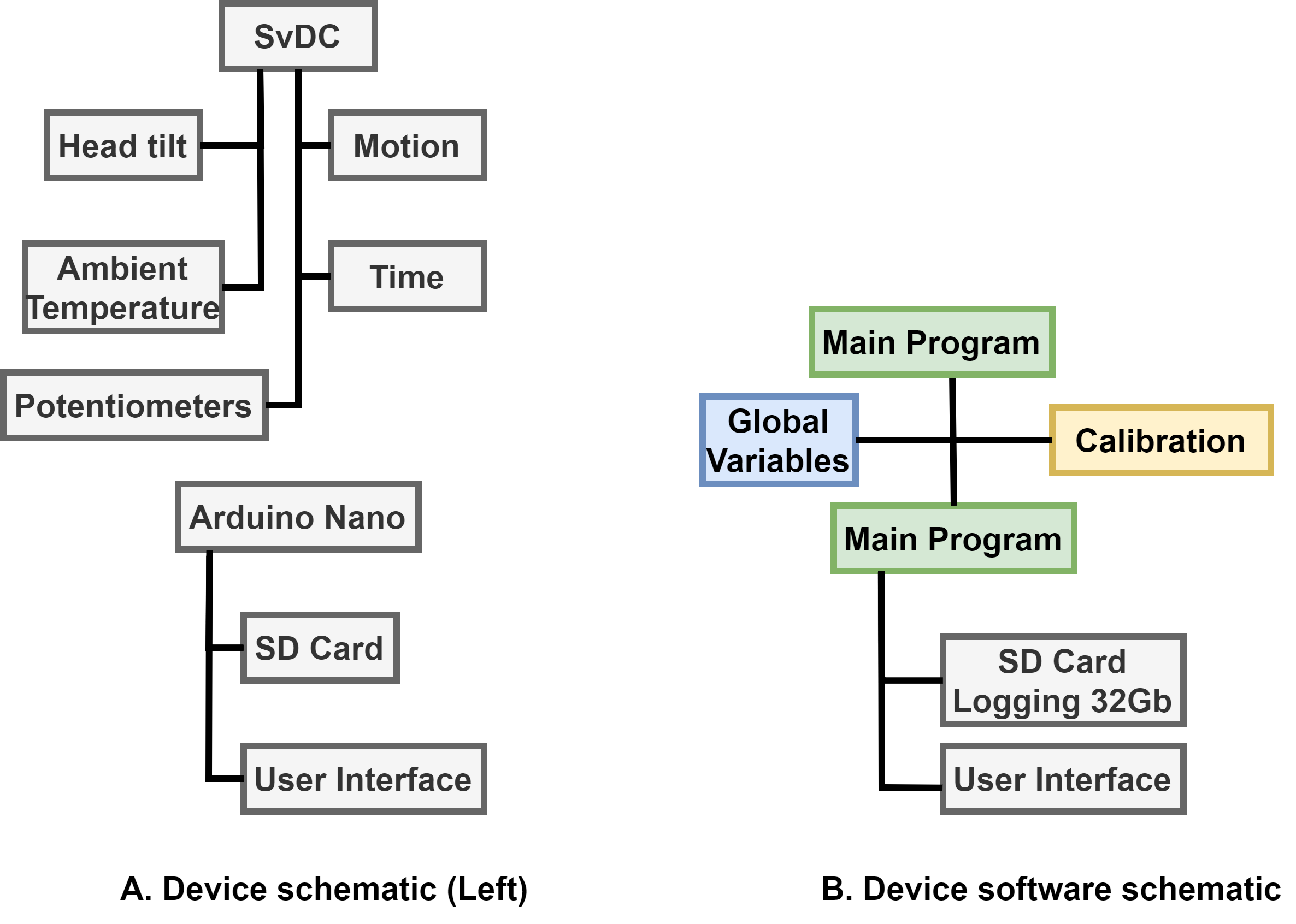}
	\caption {the proposed framework is actually a device that could be used to predict user motion-sickness in real-time using motion, head tilt, and ambient conditions \cite{Salter2019MotionSP}. Part (A) Represents the Hardware schematic and Part (B) Represents the Device Software schematic}
	\label{Motion Sickness Prediction Device Arch}
\end{figure}

The introduced model aims to capture the AV acceleration, using an Arduino microprocessor and an Inertial Measurement Unit (IMU) equipped with an OLED display. The model reads a single-user trial of over 100 samples from more than 1500 experiments to identify a subjective sickness score. Figures detailing both the motion sickness measurement device and the device software schematic are shown in Figure \ref{Motion Sickness Prediction Device Arch}. 

The proposed device also featured detecting temperature besides motion sickness signals in real-time. Furthermore, multiple journey routes were computed during the data collection and testing phases, with more than 30 minutes for each journey. The entire experiment conducted 101 trips. The samples were collected from enormous 14-seat mini-buses and passenger coaches. The authors also used data from multiple seating positions. The vehicle's air conditioning system controlled the ambient temperature. The authors also conducted testing in partly cloudy weather. 

\textcolor{black}{Overall, devices that can predict motion sickness can potentially improve comfort in AVs by enabling advanced warning to users and allowing them to take preventative measures. These devices employ a variety of sensors to track the vehicle's movements and the passenger's physiological responses to them. For instance, they may assess the user's heart rate, skin conductance, and other indicators of motion sickness. This information can then be analyzed by algorithms for predicting the motion sickness occurrence likelihood. Upon detecting a high risk of motion sickness, the device can alert the user and provide suggestions on how to prevent it. For instance, it may suggest adjusting the temperature, opening a window, changing the seating position, or taking medication. This can help take proactive measures to prevent motion sickness before it becomes severe.}

In addition to architectures that address comfort and motion sickness, there are also architectures/frameworks which address the trust aspect of autonomous vehicles. Such examples of these frameworks are discussed in the next few subsections. 

\subsection{Sense–Assess–eXplain (SAX)}
Numerous assurance and regulation critical barriers to deploy large-scale deployment of AVs were raised and discussed with their potential solutions in \cite{9304819}. The work attempts to address and clear the following objectives with the proposed approach:
\begin{enumerate}
\item Reading and analyzing weather conditions on multiple scenarios to test potential sensing modalities' limitations.
\item Continually evaluating and optimizing the perception performance and navigation methods.
\item Demonstrating the system's capability in interpreting the vehicle vision and how that impacted its decision-making.
\end{enumerate}

\begin{figure}[h]
    \centering
	\includegraphics[width=0.35\textwidth]{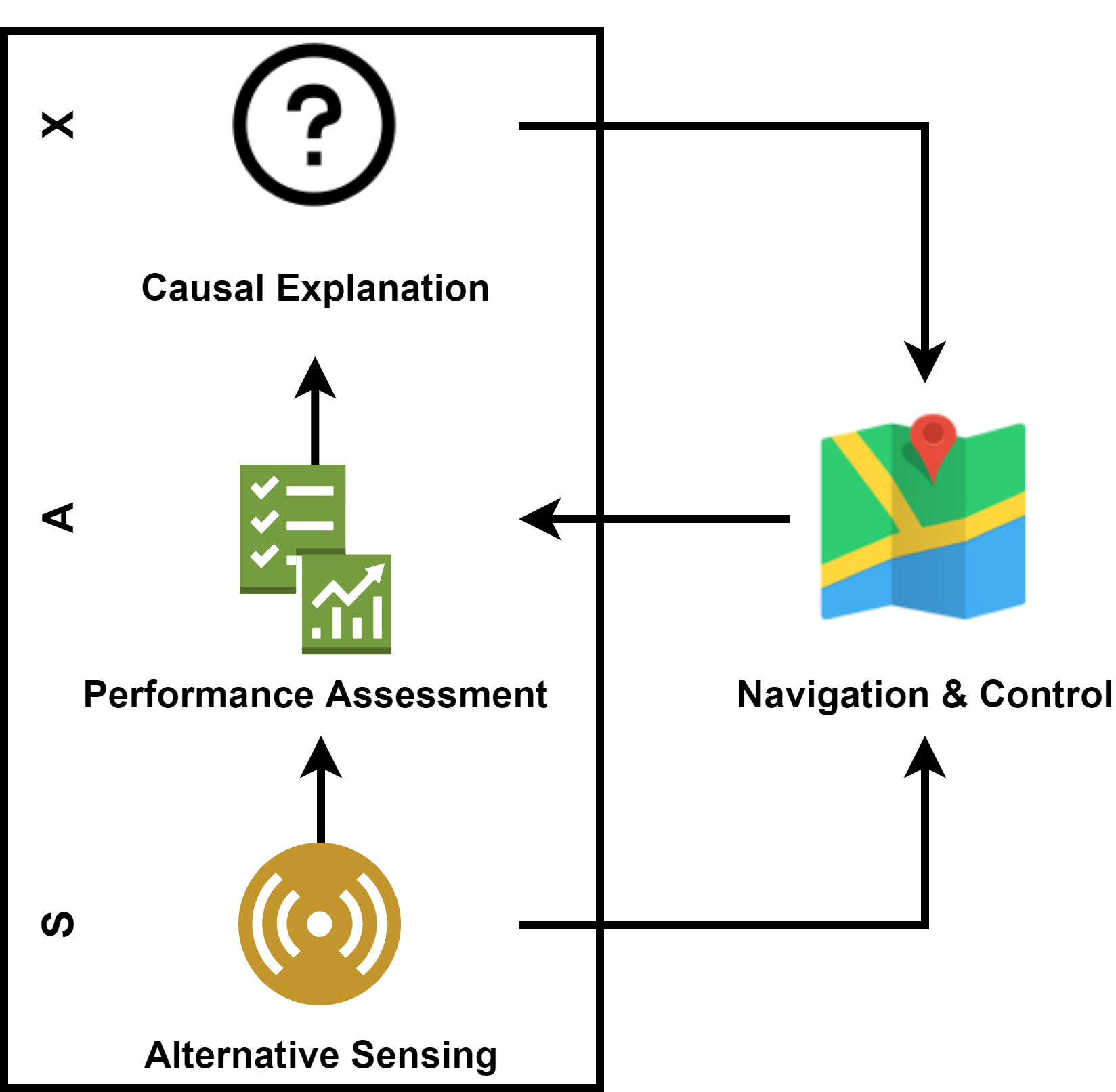}
	\caption {The developed sense-assess-eXplain (SAX) model \cite{9304819}. The proposed model is used to measure assurance and trust in AV operations.}
	\label{SAX}
\end{figure}

To satisfy these objectives, the authors develop a model known as sense-assess-eXplain (SAX). The proposed method is used to measure assurance and trust in AV operations. SAX comprises three core research aspects that aim to identify and match various levels of AV sensing and evaluation, which helps understand vehicle observations and decision-making in real-time. The AV can also clarify that the current and predicted environmental conditions impacted its performance. Figure \ref{SAX} illustrates the overall schematic of SAX.

\subsection{Real-time Trust-Building Schemes for Mitigating Malicious Behaviors}
\begin{figure*}[h]
    \centering
	\includegraphics[width=7in]{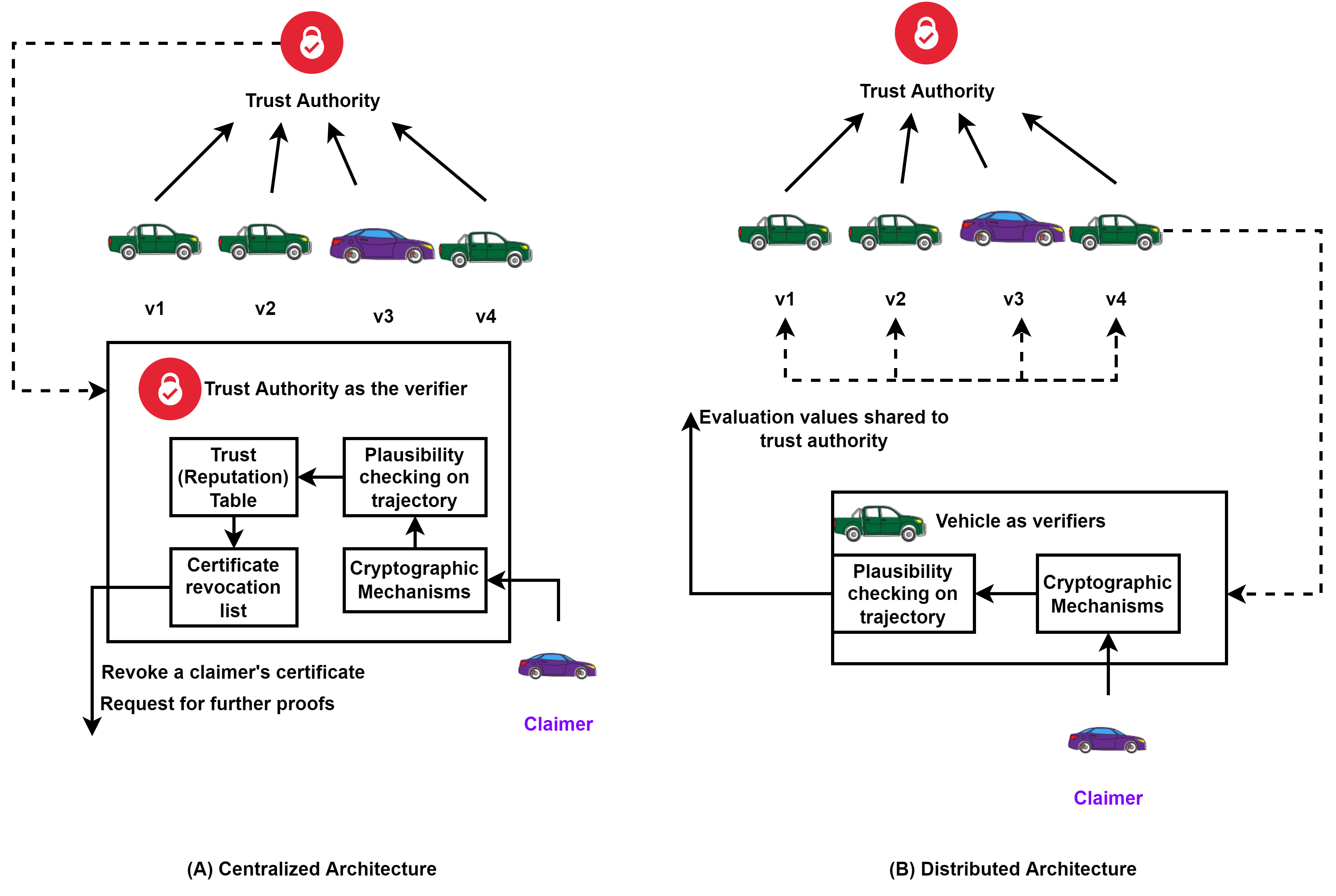}
	\caption {The two architectures developed in \cite{8917078}---(1) centralized and (2) distributed---to reduce dangerous behaviors using AVs trust measurements.}
	\label{doubletrust}
\end{figure*}

The importance of trust regarding AVs \textcolor{black}{safety and security} was discussed in \cite{8917078}, which also suggested two architectures: centralized and distributed, to reduce dangerous behaviors using AVs trust measurements. Those architectures divided the roles of the trusted authority and vehicle nodes in plausibility checking and sustaining trust to detect trust in real-time. The suggested method (shown in Figure \ref{doubletrust}) also assumes the adoption of a public-key infrastructure to facilitate inter-vehicle communications, in which case a trusted authority is in charge of issuing and revocation of certificates for certifying AV IDs.

The centralized architecture utilized the trusted authority to monitor vehicle functions and interchanges via V2X messages within a pre-defined area. Each vehicle was collected and transferred broadcasted V2X messages to the trusted authority for verification. Figure \ref{doubletrust} illustrates how the trusted authority communicates and handles the authenticated vehicle identity and message integrity and calculates trust. The capability of the authorized source to have a broader view of each AV's behaviors concerning time duration and travel distance is the main advantage of the centralized architecture to detect malicious behaviors. Nevertheless, storing trajectory and running plausibility checks of each vehicle increase memory consumption, a significant weakness of this architecture that needs to be considered when deploying an existing system in AVs.

In contrast, vehicle trajectory verification is delegated to AVs in the distributed architecture, as shown in Figure \ref{doubletrust} (B). A car verifier estimates the vehicle's location and then decides the AV trust value and shares data with the trusted authority. In that way, the trusted authority only updates its trust on the related node. This architecture's strengths are detecting local malicious behaviors that are out of vehicles' signal transmission and reducing memory consumption and associated calculations. However, this architecture's weakness is the vehicle verifier's inability to detect attackers on a global scale.

The listed work modeled a Python-based simulator for non-continued events to evaluate V2V messages exchange processes among several units. This work used centralized architecture to implement verifiers for trusted authority, while distributed architecture was developed for AVs. Moreover, four existing algorithms have been utilized in this model to test the plausibility checking module. The open-source dataset called VeReMi was used to assess malicious behaviors algorithms detecting in vehicular networks. Both centralized and distributed architectures have been deployed to test and evaluate four malicious strategies under low- and medium-density states. Despite the poor performance of the listed methods, they represent a good choice to continue investigating and gain user trust during driving AVs.

\subsection{Safe and Online MPC}
The AV navigation system that uses MPC controller design was presented in \cite{8569384}, containing longitudinal and lateral controls. Besides the safety monitoring module, it also has a cost function designed to calculate human driving habits. It estimates the required time to reach unacceptable situations, including tracking performance and comfort constraints under different tracking conditions. The strength of this kind of controller is presented in its ability to guarantee smooth trajectory and acceptable performance. Moreover, a simulator has been implemented to test the proposed MPC controller under typical urban settings.

\begin{figure}[h]
    \centering
    \includegraphics[width=3.5in]{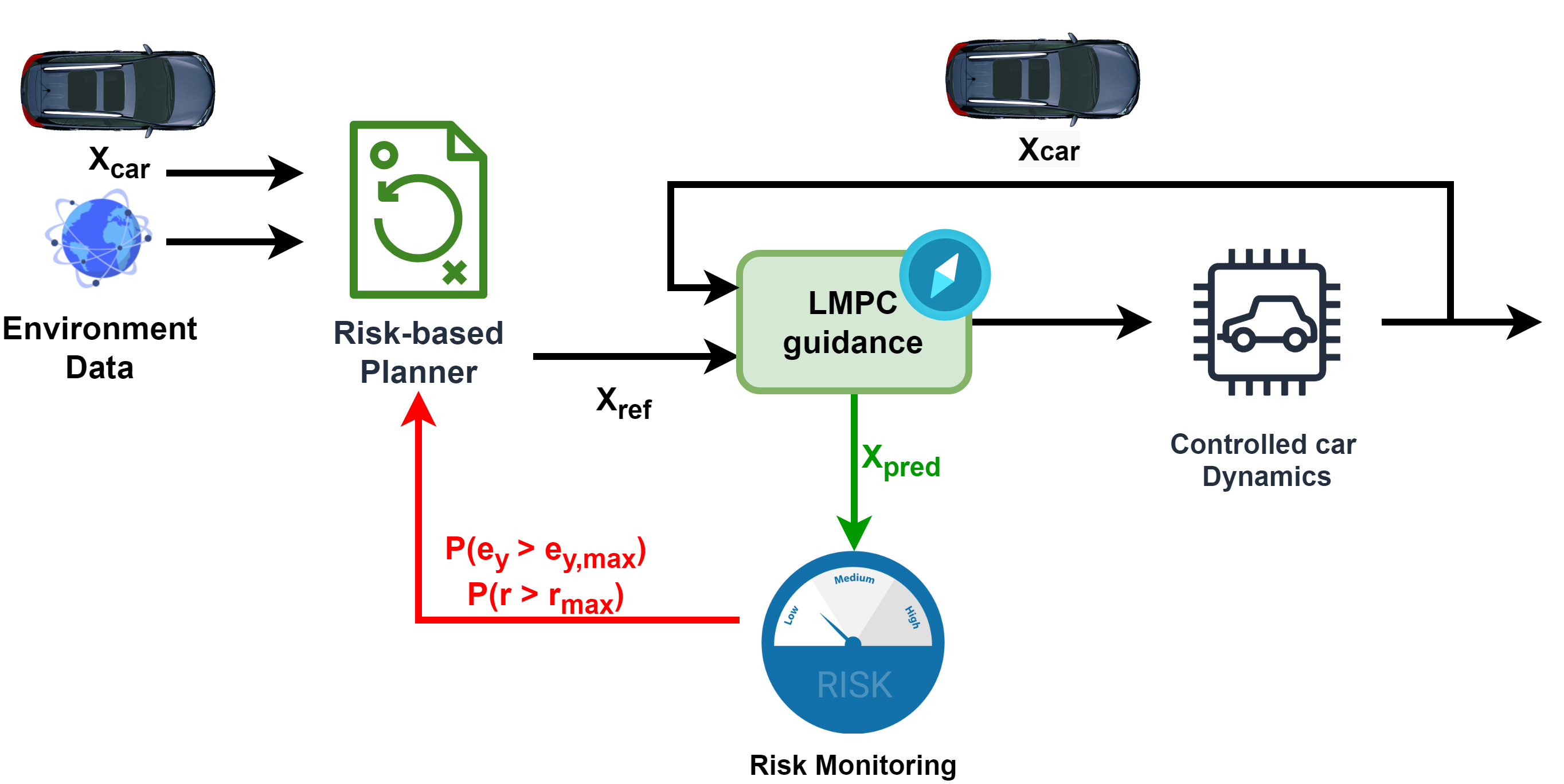}
	\caption {The proposed AV navigation system that uses MPC controller, containing longitudinal and lateral controls \cite{8569384}.}
	\label{MPC}
\end{figure}

The initial Guidance, Navigation, and Control (GNC) architecture components helped develop the suggested MPC controller to facilitate navigation and wayfinding in urban areas. The listed MPC controller controls both axes, the lateral and longitudinal. The lateral control signal manages the car steering via passing signals to a low-level controller. Calculating associated risks with the current tracking condition is another task of the MPC controller, which is considered a safety monitoring algorithm. This risk was calculated using the predicted lateral error and MPC accuracy. The proposed architecture is shown in Figure \ref{MPC}.

The suggested model was developed and examined using the 4D-Virtualize simulation engine and Robot Operating System (ROS) framework to provide semi-actual vehicle experiences. The simulation utilized IPCar, an urban electric vehicle with a maximum speed of 3 meters per second and a turning radius of 3 meters. The presented technique is robust to noise and is expected to solve ground vehicles' trajectory-tracking issues. Additionally, the proposed work also considers GPS data, making sure the GPS data is accurate as well. Furthermore, the proposed approach also achieved lower complexity. A strong benefit of this lower complexity would be that the AV gets an easier time planning its destination/route so that the passenger discomfort is potentially minimized. 

\section{Intelligent Control Methods \textcolor{black}{and Optimizations}}\label{Intelligent}
\begin {figure*} [htbp]
\centering
\includegraphics[width=.8\textwidth]{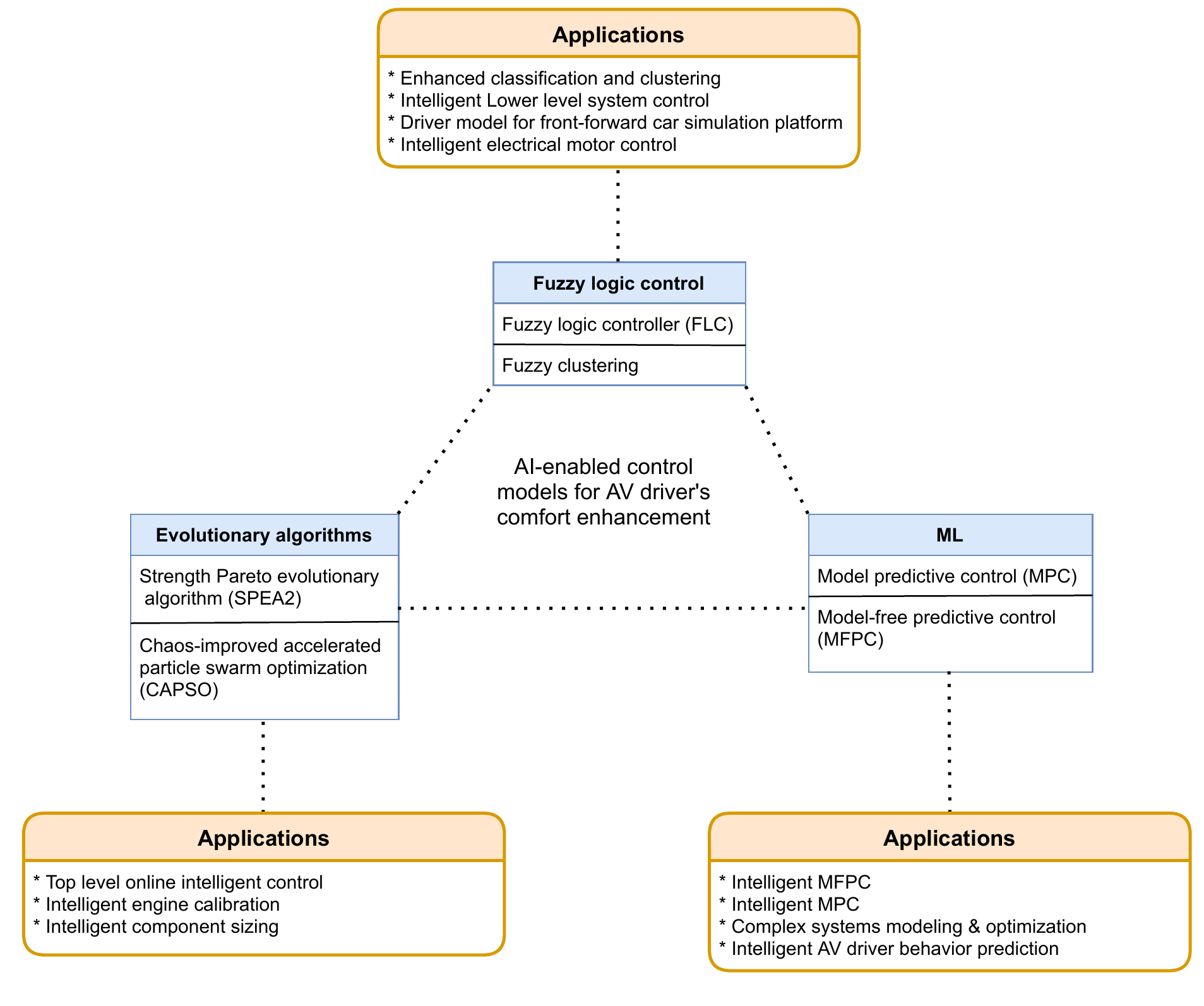}
\caption{The potential integration of AI into control models (fuzzy logic, evolutionary algorithms, and ML) with a summary of optimization applications for driving comfort enhancement \cite{samak2020control}.}
\label{AI-control}
\end {figure*}

When working for control process models and robust machine processing implementations for enhancing comfort and trust for AVs, intelligent control methods contrasted for conventional control methods are largely expressed. These include model-driven control systems, neural network management and deep learning methods \cite{chang2019using}, \cite{zhang2020software}, \cite{chuprov2020reputation}. Such algorithms are currently being used slowly in widely used automotive regulation. Figure \ref{AI-control} presents an overall taxonomy of the application of these intelligent control methods.

\subsection{Model-based Control} 

Model-based control is also known as Model Predictive Control (MPC), Rolling Horizon Control (MHC), and Receding Horizon Control (RHC). It is a model prediction-based computerized control system, which has been extensively researched and applied over the last few years \cite{garcia1989model} \cite{rossiter2003model}. The basic theory can be summarized as: an open-loop optimization problem of the final time domain is overcome at each sampling point and the first dimension of the control sequence obtained is implemented on the controlled object according to the current measurement information currently obtained. Invoke the above procedure in a sampling instance and then restart and address the question of optimization with new calculated values. The key distinction between predictive model management and conventional control methods is to solve open-loop problems digitally in order to achieve open-loop optimization sequences. The predictive monitoring algorithm consists primarily of four parts: predictive model, adjustment of feedback, rolling optimization, and comparison trajectory. The first (or first part) aspect of the optimization approach is better added to the method.

\subsection{Neural Network and Imitation Learning Control}

Neural control is the study and usage of certain structural mechanisms of the human brain and the control of the system by human knowledge and experience \cite{ge2013stable}. The control problem can be considered to be a pattern recognition problem using neural networks, and the mode defined is a ``shift" signal which is transformed into a ``compassionate" signal. The most remarkable feature of neural control is its ability to learn and adapt. It is achieved by constantly correcting the weight connections between neurons and storing them discretely in the connection network. It has a good effect on the control of nonlinear systems and systems that are difficult to model. In general, neural networks are used in control systems in two ways: one is to use them for modeling, mainly using neural networks to arbitrarily approximate any continuous function and the advantages of its learning algorithm, there are two feedforward neural networks and recurrent neural networks One type; the other is used directly as a controller.

The vehicle's longitudinal dynamics are handled via a pre-tuned PID controller, whereas the lateral dynamics are handled using an end-to-end trained neural network. The main advantages of deploying imitation learning-based control in autonomous vehicles are its straightforward implementation in addition to its quick and fair training capabilities, i.e., in comparison with reinforcement learning. Furthermore, the neural network facilitates the driving learning process for vehicles based on a provided dataset. Although this can be accomplished without complex formulations governing the system's input-output relation, biased and/or improperly labeled data may lead to incorrect learning by the neural network. Therefore, advancements are still needed in order to attain reliable performance and accuracy when adopting such an imitation learning-based control for autonomous driving \cite{samak2020control}.

\subsection{Deep Learning Control}

The research of Neural Networks, meaning deep neural networks (DNN), aka artificial neural networks (ANN), results in profound thinking \cite{xiong2016combining}. This model's strength is presented in the ability to automatically select features of high-dimensionality data, which helps avoid complex feature manual selections. The most important feature of using deep learning in such fields is handling feature extraction techniques within the model (without using separate data preprocessing techniques to detect features) and fitting a model. For control systems with high-dimensional data, introducing deep learning has a lot of significance. Lately, some examinations have applied deep learning approaches to control systems, such as deep belief network (DBM) and Stacked Automatic Coding (SAE); CNN \cite{masaki2012vision}; and Recurrent Neural Networks (RNN).

\subsection{Reinforcement Learning Control}

Unlike imitation learning-based control (neural network methods), where the learner that is trained to mimic the trainer's behavior may never outperform the trainer's performance capability, the reinforcement learning method learns through trial and error. The ``agent" trained using reinforcement learning discovers the surrounding environment and independently explores new control strategies. Specifically, $\pi$, the policy to be trained for optimal control can be a neural network that behaves as a function approximator controlling the relation between actions (outputs) a and observations (inputs), $\pi(a|o)$. The agent is rewarded upon carrying out valid actions while penalized for any improper actions, and thus the goal of the agent, given a specific policy, is to maximize the expected reward per $\argmax{E} [R|\pi(a|o)]$. However, defining the expected reward is still highly critical as rewarding the agent for carrying out good actions may enable them to randomly wander pursuing the reward maximization (i.e., risk of never accomplishing the training objective) \cite{samak2020control}.

Last, a summary of comparisons between the traditional and intelligent control methods is provided in Table \ref{comparison_controlMethods}. The potential integration of AI into control models (fuzzy logic, evolutionary algorithms, and ML) with a summary of optimization applications for driving comfort enhancement is further depicted in Figure \ref{AI-control}.

\begin{table*}[htbp]
\centering
\caption{Traditional vs. Modern Control Methods.}
\label{comparison_controlMethods}
\begin{tabular}{p{1.1cm}<{\centering}p{2cm}<{\centering}p{4.9cm}<{\centering}p{2.8cm}<{\centering}p{4.8cm}<{\centering}}
\hline  
\textbf{Control}&\textbf{Method}&\textbf{Pros}&\textbf{Cons}&\textbf{Function}\\
\hline  
\textbf{Traditional} & PID \newline Fuzzy \newline Optimal \newline Sliding mode & Widely used; Increases system performance and simulation controlling output is easy to achieve (i.e., Fuzzy); Better robustness, adaptability and fault tolerance (i.e., Fuzzy)  & Designed for mostly linear systems (i.e., PID), which do not change complex characteristics over time & Focuses on the study of the conditions and methods to optimize the control system indicators (i.e., Optimal), stability, and comfort (e.g., steering); The controller switches constantly in a hopping fashion according to the current system condition (i.e., sliding mode) \\
\hline 
\textbf{Intelligent} & Model-based \newline Neural Network \newline Deep learning & Control systems with high-dimensional data; Efficient feature extraction and model fitting; The ability to learn by continuously correcting the connection weights between neurons and storing them discretely in the connection network; Provides a good effect on the control of nonlinear systems and systems that are difficult to model & Boundary detection; Semantic segmentation; High computation power; Requirement of a large amount of training data; Instability under adversarial perturbations & The feature selection process is done automatically, which significantly expedites the process, especially when dealing with big and high-dimensional data; Enables control process models and robust machine processing implementations for an enhanced driving experience; Enhance driving experience/comfort based on driver's behavior prediction \\
\hline
\end{tabular}
\end{table*}


\subsection{Trajectory Optimization}
\subsubsection{Linear Quadratic Path Planning}

In traditional driving, if the driver feels uncomfortable, he can easily adjust the driving style to improve the comfort in driving and thus the driving experience. This role was passed to the car operating system after the vehicle implemented autonomous functions. A system for determining a driving efficiency is required in order to achieve safe riding in autonomous vehicles.

Considering the importance of ride comfort measurement during the development and adjustment of autonomous vehicles, the Svensson team designed a system of adjustment strategies including frequency analysis \cite{svensson2015tuning}. The system evaluates YSIS and ride quality by simulating the lateral control system. Therefore, they updated the controller configuration and suggested H as the frequencies weighted linear quadratic controller in order to further boost the conduction efficiency output through the route planning algorithm. By measuring and testing their frequency range, the frequency quality of the acceleration data is calculated. The system chooses to use the power spectral density (PSD).

The system is a lateral control system to change the lateral route preparation for self-supporting vehicles, taking into consideration ride quality efficiency \cite{svensson2015tuning}. This works by developing a way of determining the right set of tuning parameters by looking at the balance between reaction and riding standard. These sums can be measured by graphical instruments \cite{svensson2015tuning}.

\subsubsection{Path Planning and Optimization}
The method based on trajectory optimization is a widely used method in path planning \cite{ziegler2014trajectory, liu2015predictive}, which represents route planning as a question of optimization, taking into account the appropriate vehicle efficiency and related restrictions \cite{amer2017modelling}. Its main advantage is that it can flexibly and easily encode various requirements on the required path of the AV. At the same time, advances in solvers for online optimization (CVX \cite{grant2008graph} and Ipopt) have helped to provide real-time, fast, and reliable path generation. Owing to the normal complex and unpredictable driving conditions (which can not be completely modeled), the model predictive control (MPC) approach is also used for online AV track preparation \cite{gao2012spatial, beal2012model, carvalho2013predictive}. MPC solves a variety of common problems in the optimization of finite times and during the planning process can change the environmental status.

The Liu team suggested an MPC-based path plan system, which tackles multiple modes in a single sense \cite{liu2017path}. They do not use preset rules to determine some operations, but rather to resolve online non-linear MPC issues. The paths created by MPC are automatically determined by operations including lane shifts, lane-keeping habits, ramp merges, and intersections. They have chosen the relaxed convex method for determining lane change and line-keeping activities in the objective function of MPC \cite{liu2017path}. Around the same time, cars were designed as polygons in order to ensure driving safety. A form of restriction has been created to implement avoidance of collisions between auto and surrounding cars. They also have a lane-related area that is designed for the easy and efficient handling of nearby vehicles in the same lane \cite{liu2017path}.

\subsection{Comfort Optimization Techniques in AVs}
\subsubsection{Overall Comfort Optimization}

Comfort is principally affected by the following two factors: \textit{Jerk}, which refers to the acceleration shift rate; and \textit{Bending Rate}, which refers to the change in curvature. Higher comfort means a smaller rate of change in acceleration and a smaller rate of change in curvature. The corresponding cost function can be set as ``acceleration + acceleration change rate + curvature + curvature change rate''. Therefore, the optimization of comfort is mainly related to horizontal optimization, vertical optimization, and midline issues.

\subsubsection{Response Time Optimization}
The reaction time is a system response time, which involves time for cloud network monitoring and the analysis of laboratory meetings and the system measurement, and the breaking time for the car. The system's overall response time can not be greater than 10 milliseconds if the braking distance of 100 km is not to surpass 30 cm, and the individual reaction speed of the strongest F1 conductor is around 100 milliseconds. 

Thousands of self-directed sensors will increase their speed and intelligence. Such devices produce unimaginable data sizes well in advance of any other IoT device. If data needs to be interpreted and analyzed faster than in current 4G systems, self-driving vehicle systems require exceptional data processing and speed to simulate time for human reaction. 5G will speed up the network and reduce the latency of self-propelled vehicles in order to achieve faster response times.

Leading semi-producers worldwide, including Intel and Qualcomm are going for the ASIC revolution to tackle the difficult question of integrating 5G bandwidth with wireless radio and antenna architecture \cite{ahmed2018survey}. Briefly, these firms create chips to transform autonomous cars into mobile data centers, allowing cars to make complex decisions in real-time.

In terms of supercomputing power, along with the continuous improvement of cloud computing and on-board computer computing power, the on-board computer system can process more complex tasks in a shorter time and realize automatic driving real-time perception of road conditions, intelligent decision-making, and control. With the emergence and wide application of excellent algorithms such as ML and DL, AI has entered a new stage of rapid development after 2013. Applied to the field of automatic driving. The error rate of vehicle recognition using deep learning methods is lower than that of traditional methods adopted in the year.

Low end-to-end latency is more dependent on improvements in sensors, processors, algorithms, and machine transmission. After the deployment of ultra-low latency 5G networks, we believe that more advanced communication technologies will bring more innovations to car companies and make AVs more secure.

The real innovation of the new system lies in its comprehensive signal processing capability. This allows all processing to be performed directly within the module, and the system selectively filters the data from the radar system and the stereo camera so that the processing can be performed immediately or deliberately delayed to subsequent processing stages. Identify irrelevant information, but do not forward it. Sensor fusion is applied to data fusion between the camera and radar. Then, the neural network evaluates the data and determines the actual traffic impact based on machine learning techniques. Therefore, the system does not need to send status information to the vehicle, but only needs to send a response command. This frees the vehicle's bus to process important signals, such as detecting a child suddenly running on the road.

\subsubsection{Motion Sickness Optimization}
To optimize the problem of motion sickness in autonomous driving, researchers have made a lot of efforts, summarized as follows.
\paragraph{Anti motion sickness glasses frame}
French automaker Citroen has designed an anti-motion sickness eyeglass frame called SEETROEN \cite{meschtscherjakov2019bubble}. In this solution, there are blue liquids inside the four rings. These liquids are located on the ``visual edge" of the wearer. When the vehicle turns, brakes, or bumps, the blue liquid will also flow with it, so that the visual information is obtained by the eyes. Consistent with the movement signals felt by the cochlea, keeping the brain's perception consistent. It is understood that SEETROEN can alleviate motion sickness within 10 minutes, and the success rate is as high as 95\%.

\paragraph{Trick the brain}
One strategy is to trick the passengers' brains into thinking they see the movement of the car, even if they do not. The University of Michigan Transportation Research Institute has patented a system that simulates visual cues that people get when viewing from outside a car \cite{sivak2018universal}. This technology mimics the vestibular input corresponding to the vehicle's speed, acceleration, yaw, and lateral motion. Its goal is to blind your sight to the vestibular system. This allows passengers to focus on other things, which has been a big selling point for autonomous vehicle technology since its inception, without the effects of motion sickness.

\paragraph{Smooth driving and motion sickness warning system}
Google's Waymo has filed a patent describing a system that eliminates motion sickness by identifying a route that minimizes motion sickness \cite{larner2018method}. Waymo's patent solution to motion sickness is to calculate the motion sickness index of each route, and finally select the route with the smallest motion sickness index. And communicate motion sickness information with passengers through the display panel.

The system will suggest providing less complicated driving roads to prevent motion sickness, or send passengers an alert on bumpy roads prone to motion sickness, and recommend that passengers do not look down or read books during the journey \cite{larner2018method}. There are also special seats in the car for people who are prone to motion sickness, and the algorithm evaluates the congestion of alternative paths in advance to determine the motion sickness index. Waymo's technology will study the car's back-and-forth acceleration and sway to calculate whether motion sickness may occur on a particular route. When passengers specify that they feel unwell, the system can also change their driving style or route. In addition, Waymo unmanned vehicles can adjust driving styles based on passenger motion sickness reactions, such as increasing the distance from the car in front.

\paragraph{Sensory simulation system}
Uber plans to use vibration and mobile seats, airflow targeted at the face or other parts of the body, and light bars and screens to prevent passengers from feeling uncomfortable. It applied for a ``sensory stimulation system" that is essentially a vestibular system that trains passengers so that what they see is related to their feelings \cite{sweeney2017sensory}. One mechanism proposed for this includes a light bar that surrounds a portion of the interior and provides visual cues to alert passengers to acceleration, braking, or changes in direction. It may use color and brightness levels to indicate different actions.

Uber not only attempts to unify visual and cochlear information, but also provides a set of immersive experiences including vision, hearing, seating, and airflow. You might also hear the breeze rushing through the emergency stop as the vehicle brakes abruptly \cite {sweeney2017sensory}. First, visually, Uber's car has a set of AR (augmented reality, overlaying virtual information in the real environment) system, which can use the environmental information outside the car captured in the car to present in the car, making up for the visually immobile perception Dislocation to synchronize somatosensory information. A matching tactile system is also provided on the seat, which will give the corresponding airflow when turning sharply and rapidly.

\paragraph{VR games}
Holoride is trying to eliminate motion sickness with VR glasses \cite{fereydooni2022incorporating}. The movement in VR games is consistent with the vehicle trajectory so that the movement we see is consistent with the movement felt by the body.

\begin {figure} [htbp]
\centering
\includegraphics [width=\columnwidth] {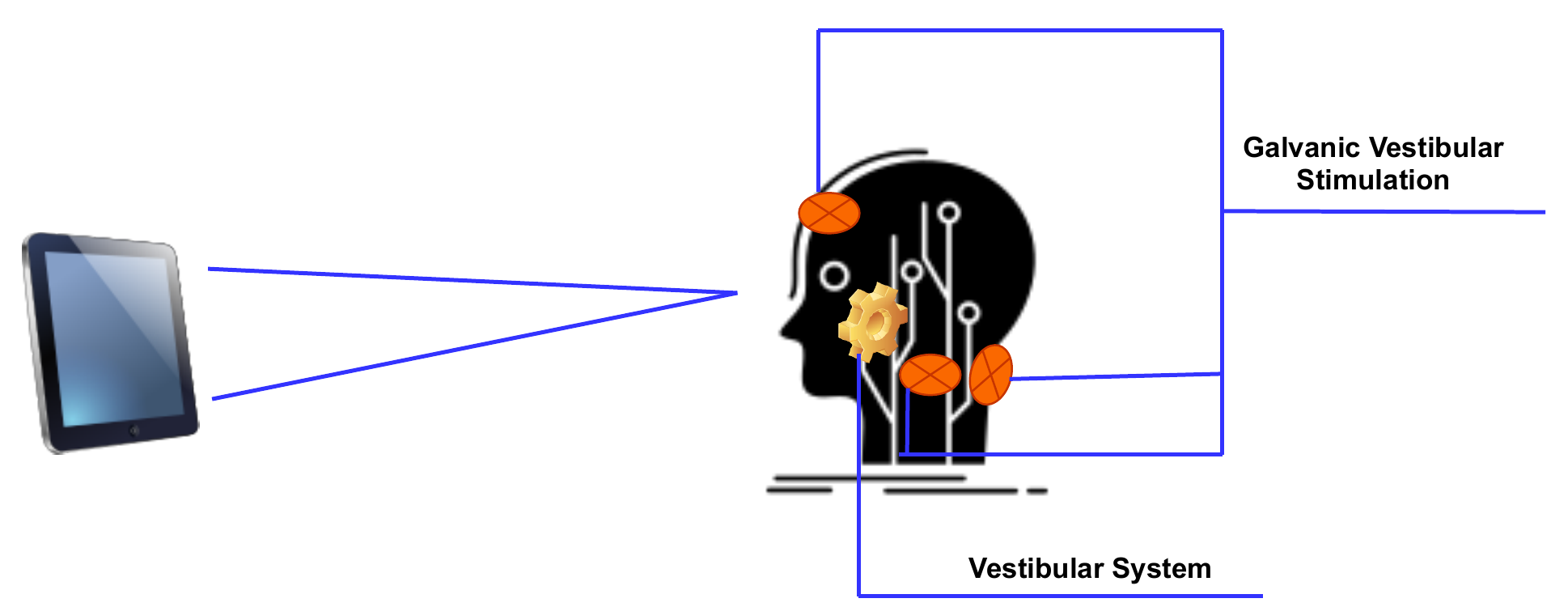}
\caption {vMocion's 3v$^{TM}$ Platform based on GVS to convert motion feedback to users using 4 electrodes in the ears, neck, and forehead \cite{anastasijevic2016mayo}.}
\label{vMocion}
\end {figure}

\paragraph{Vestibular electrical stimulation}
Most of the above methods change the visual information so that the visual information is consistent. vMocion attempts to alter vestibular information to align it with visual information \cite{sergeeva2017virtual}. vMocion's 3v$^{TM}$ platform is a device based on electrical vestibular stimulation (GVS), which can potentially be used to convert motion feedback to users \cite{anastasijevic2016mayo}. By placing 4 electrodes in the ears, neck, and forehead, electrical stimulation is used to align the movement detected by the nervous system with the visual content. Figure \ref{vMocion} shows vMocion's 3v$^{TM}$ platform.

\par
Samsung has already created an Entrim 4D headset built on the GVS device. The team sees Entrim 4D as a potential response to the movement disorder issue \cite{Samsung}. However, the technology is still under development.

\subsubsection{Lane Change Maneuver and Optimization}

The required speed can be regulated according to the route radius and a quick transformation can be achieved \cite{kilincc2012determination}, This eliminates the disparity between vision and human vestibular system in fluid travel, making the journey smoother. Smooth motoring techniques will boost vehicle satisfaction and closely track accelerating, braking and steering. The quick lane changes or curves generate a vehicle's lateral acceleration, so developing new sufficient lane change maneuver algorithms helps improve AV comfort.

There are different techniques for modifying the routes of driverless vehicles, with Bezier curves the most widely used way to create routes for lightweight communication and marshaling (LCM). A Bezier curve-based route planning algorithm has been proposed to create a reference path to meet the roadside restrictions in \cite {choi2008path}. Also, a Bezier curve-based collision-prevention technique was introduced by \cite{han2010bezier} that considers all path lengths between automobiles. Chen et al. considered the incomplete constraints and yaw rate parameters of vehicle steering and developed a Bezier-based lane change algorithm \cite{chen2013lane}. In another work, Kawabata et al. suggested to Bezier a 3-dimensional refer to a Curve generation constraint optimization method \cite{kawabata2015path}. Moreover, an efficient LCM Bezier curve-based algorithm was developed \cite{bae2019lane} to estimate potential lateral acceleration to be employed in a local road planning phase to transfer the side motion of the vehicle during lane change maneuvers smoothly. The algorithm measures the curvature of the Bezier curve by the parameters of the lateral acceleration factor reached by the operator in order to restrict lateral motion. If lane change maneuvers are required, space availability must first be confirmed to generate a local path. The planning block then produces alternative pathways and tests the local pathways generated to determine whether they fulfill a certain acceleration limit.

\section{Applications of Driving Control Methods and Comfort Optimization Techniques for AVs} \label{Apps}
This section covers some applications of driving control methods and comfort optimization strategies for AVs in various scenarios. Interestingly, a couple of these works incorporate path-planning mechanisms, as seen in \cite{rokonuzzaman2021review} and \cite{abdallaoui2022thorough}.

\subsection{Speed Control Framework for AVs - Rough Pavements}

A deep reinforcement learning-based speed control strategy has been proposed for AVs on rigid pavements \cite{du2022comfortable}. The developed approach aims to estimate riding comfort and energy-efficient speed control. The proposed framework utilized open-access real-world data for training the model. The model achieved energy and computation efficiencies by 8\%, 24\%, and 94\%, for which the authors claimed their model would be a good choice for controlling AVs speed on suburban roads. 

\subsection{SAINT-ACC}

An adaptive deep reinforcement learning-based cruise control system called Safety Aware Intelligent ACC (SAINT-ACC) has been proposed to optimize the efficiency of traffic, safety, and comfort \cite{das2021saint}. SAINT-ACC can perform driving safety assessments via modification of safety model parameters in response to various traffic conditions. Moreover, the proposed approach also utilized a dual reinforcement learning system, improving traffic efficiency, safety, and comfort. A separate reinforcement learning component is designed to find a specific threshold based on traffic data obtained from the environment. This model interacts with another reinforcement learning-based model to identify the insufficiency in the riding flow, safety, and comfort. The two reinforcement learning agents are trained via numerous highway scenarios, including those with on-ramp and off-ramp highways. A simulation platform is also used to perform experiments over 12,000 runs for training. Results obtained from these tests confirmed SAINT-ACC's ability to enhance driving safety, traffic flow, and comfort.

\subsection{Real-time NMPC Path Tracker}

A non-linear predictive-based path tracker model for AVs has been proposed \cite{farag2021real} using C++ and several powerful libraries to guarantee optimal real-time performance. The listed work claimed it featured by high flexibility in a way that allows the incorporation of multiple objective terms to form a cost function. The proposed model aims to provide precise tracking while maintaining a comfortable lift. After testing the proposed approach on numerous simulation studies that featured intricate tracks with many sharp turns, the authors confirmed that the proposed path tracker could outperform a regular PID-based controller of an AV.

\subsection{TDR-OBCA}

An optimized collision avoidance model for AVs driving has been presented in \cite{he2021tdr} to drive comfortably and efficiently. The TRD-OBCA is designed to provide a smooth collision-free trajectory model for AVs in accessible environments, such as parking lots or off-road areas. The TDR-OBCA algorithm has been tested and validated in simulations and real-world road experiments. Valet parking and various obstacles have been illustrated in simulations. This approach claimed to drastically reduce the loss by 97\%, while increasing the computation time by 45\% and improving the overall driving comfort scores. The proposed model is claimed to lower the steering control output by more than 13.5\%, for which the authors suggested using it for real-world AV applications.

\subsection{Model Predictive Control}

The purpose of \cite{luciani2020model} was to propose a method for designing a model predictive control scheme to enhance the passenger's comfort for AVs. According to the authors, the proposed strategy is different from previous attempts because of its wide applicability in various driving scenarios and incorporates an offline method weighting parameters of the model predictive control scheme. The authors tested the proposed approach with experiments conducted in Matlab/Simulink and the tests were done using the highway and urban setting scenarios. The proposed approach utilized the MSDV and acceleration coefficient metrics to evaluate the model performance to meet the ISO 2631 standard requirements. The system also considers the ride comfort and drive quality described by ISO 5805 standards. Ride comfort is identified as ``a subjective state of welling and absence of mechanical disturbance concerning the environment and ride quality."

\section{Implications, Insights, and Future Works} \label{insights}

\subsection{Comfort and Health Risks}

AVs may strengthen the disadvantaged position of humans in the battle against pollution and sedentary behavior, affecting the health of drivers and increasing health risks. Automated driving reduces traffic pressure and the apparent danger of car travel, which leads people to rely more on vehicles. Even for short distances, people may choose to drive by car instead of walking.

Sedentary motion and frequent exposure to pollution are the two most common risk factors for heart disease. Statistics show that about one-quarter of Americans die of heart disease each year. People who drive frequently often live a sedentary life, and their incidence of obesity, diabetes, and heart disease will also increase.

In the 20th century, more and more people owned their own cars, which allowed them to work in the city without having to live in the city, which means that the core of the city is surrounded by a growing number of suburban residents. Therefore, for many people, walking is no longer an option. The growth of the city also presents significant problems for public transport: the need for buses or subways with longer kilometers is growing as the number of residents grows. This situation requires everyone to make a choice, either buy a car or bow to reality and choose a longer commute.

A future full of self-driving cars means less walking time and more urban expansion for people, and if the development and progress of electric vehicles cannot keep up with the requirements of the times, then pollution is bound to increase sharply. After all, when people drive self-driving cars and can do anything they want to do on the commute, the driver doesn't care whether the time on the road is 20 minutes or an hour.


\subsection{Benefits and Costs of Driving Comfort Adoptions}

\subsubsection{Practices in Designing Safety and Comfort of AVs}
The vehicle must not completely rely on the GPS function. The vehicle needs to cross-reference the pre-drawn map and real-time sensor data, and after comparing the two, determine the location of the vehicle. A notable example of such a design practice is the Google Waymo use case. The Waymo team aims to develop a comprehensive high-resolution 3D map before driving, which will show road lines, curbs and sidewalks, streets, crosswalks, stop signs and other road conditions \cite{waymo2017road}. Furthermore, the self-driving car sensors and software must continuously scan the surrounding environmental information within 300 meters and read traffic control information from the color of traffic lights and railway crossings Temporary Stop Sign. To enhance driving comfort levels, the deployed algorithms will forecast the potential travel on the basis of the present speed and direction of various complex path goals, and can even distinguish the difference between vehicles and cyclists or pedestrians. The software needs to use information obtained from other road users to anticipate a variety of possible driving routes and predict how the road surface will affect other targets around the vehicle.

\subsubsection{Benefits}
Due to the reduced time, energy, and money required to travel, GTC is expected to decrease. The comfortable travel, first of all, the safety of travel, reliability of travel time and other activities than driving like work, meetings, food or sleep, etc. are more important \cite{milakis2017policy}.

The effect of self-employed cars on the time demand of travelers has been minimal. A preference survey conducted by Yap, Correia and van Arem in the Netherlands found that the use of fully automatic (level 5) has a higher time value than the use of manually driven vehicles as a train travel exit mode \cite{yap2016preferences}. Such researchers offer the impression that driving a driving vehicle will offer passengers a feeling of uneasiness, lack real-life knowledge with self-drift vehicles, and outbound trips are typically short trips and travelers do not have the ability to completely benefit from self-driving vehicles, including protection in their journey. Cyganski team suggested that in the questionnaire survey in Germany, only a very small number of respondents referred to the working ability (level 3 and higher) of autonomous vehicles as an advantage \cite{cyganski2015travel}. In comparison, most respondents accepted that behaviors, such as looking, chatting, or enjoying music, should also be relevant while traveling in a self-employed vehicle which they usually practiced when driving a regular car. Answers who have been found working in the current route prefer to work in independent vehicles.

The free passage capability, distribution of vehicles along the roads, and consistency of traffic can be advantageous for individual vehicles \cite{milakis2017policy}. Higher freeflow levels and reduced traffic levels will improve road capacity and thus decrease congestion time, which is the reduction in the amount of queued pollutant emissions. Quoted congestion delays are minimized as a result of improved road availability and minimized (or even eliminated) duration of parking search due to car parking space. The use of modern cars has also expanded and can take less time to drive.

In their study, Hoogendoorn et al. concluded that self-driving would decrease traffic congestion by 50 percent, and this degree of reduction can be improved with the help of technology in cars and buses \cite{hoogendoorn2014automated}. The Michael team revealed that, although the degree of coordination between cars is growing, efficiency is being seen in the simulation of a single-wave automatic highway network \cite{michael1998capacity}.

Autonomous vehicles should be implemented to facilitate the growth of motorway and car-sharing systems. The running costs for ride-sharing and car rental systems will be greatly lowered for autonomous vehicles. Autonomous cars can fulfill individual travel criteria easily at lower costs and greater versatility \cite{milakis2017policy}. Then city residents may agree to limit or even eliminate the number of cars they own. In the hypothetical and real town of Zurich, Zhang et al showed simulated in Switzerland that about ten and fourteen traditional cars can be replaced by each shared self-driving car \cite{zhang2015exploring}. The substitution rates for private vehicles, according to Chen, Kockelman, and Hanna, are between 3.7 and 6.8 if automotive charging is not used in the public, hybrid and self-driving cars \cite{chen2016operations}.

Regional and local area preparation can be influenced by autonomous vehicles. People can choose to use self-driving shared vehicles instead of driving themselves, so parking areas can be reduced, and the increased space can be used to build more suitable public facilities for life. Reducing parking demand can also bring about changes in land use and architectural design.

In the background of the urban economy, Zahalenko studied the effect of fully autonomous vehicles \cite{zakharenko2016self}. The researcher has developed a semi-circular single-center two-dimensional city model that has been calibrated to representative cities in the United States. He believes that workers can choose between not commuting, traditional commuting, and automatic commuting, taking into account the variability of each choice, parking fee, and fixed charges. Statistics suggest that about 97 percent of daily car parking needs will be moved beyond the city center to ``dedicated parking areas." In addition, the economic growth in the city center will have a favorable impact and would raise land prices by 34\%. On the other hand, the decreased travel costs arising from the use of electric vehicles will lead to urban growth, and land rent will decline by about 40\% from areas beyond the city center.

Due to the lower risk of accidents and the higher fuel efficiency of vehicles, the efficiency of traffic flow is improved and the monetary cost of travel can also be reduced. The limited distance allows air resistance to be minimized and therefore the consumption of fuel and costs to be decreased. The implementation of auto drives will also offer incentives for electricity and pollution.

Wu et al. showed a fuel efficiency optimization program that can suggest the right acceleration/deceleration values for the driver or automatic device, considering vehicle velocity and acceleration, as well as existing speed limit, size, road lighting, and road sign \cite{wu2011fuel}. Their studies with driving simulators in urban environments of signalized intersections have demonstrated a 31\% drop in drivers' fuel consumption.

\subsubsection{Costs}

It is expected that autonomous driving incurs additional costs where the AV cost (e.g., vehicle equipment development, equipment maintenance, etc.) is regarded as the main constraint to achieving the maximum level of benefits. The overall AV cost comprises manufacturing cost, operating cost, and social cost.

\paragraph{Manufacturing cost}
\begin{itemize}
    \item \textit{R\&D}: Needless to say, R\&D is the core of autonomous driving. Because it is difficult to achieve the desired output in recent years, the unit R\&D cost will be high. If a self-driving R\&D team has 500 people and the average annual salary is 200,000 US dollars, the total salary the company pays to employees every year is 100 Million. If this company has an amazing production capacity, it can build 5,000 self-driving cars every year, and the unit R\&D cost is \$ 20,000.
    \item \textit{Sensor}: Sensors are the most important part of the cost of autonomous vehicles \cite{fazlollahtabar2019cost}. Taking General Cruise as an example, in November 2017, a Cruise will be equipped with 14 cameras, 5 LiDARs, and 21 radars. Moreover, the sensor technology is updated very quickly, and the LiDAR introduced at the beginning of the year may become obsolete by the end of the year. So even if the cost of the sensor itself is getting lower and lower, the cost of replacement will not be reduced. 
    \item \textit{Assembly and mould}: The sensors of self-driving cars are many and complicated, and the assembly process will not be simpler than traditional cars. A set of molds for traditional cars can reach 200 Million US dollars, if it is a new design it will reach 600 Million US dollars. For self-driving cars, if the body is a new design, and the replacement speed is fast, the mold cost will remain high.
    \item \textit{Materials}: Most self-driving car R\&D companies have begun to use energy-saving and environmentally friendly materials to manufacture car bodies to ensure that the car body is light and energy-saving \cite{fazlollahtabar2019cost}. Waymo uses plastic as the windshield and fiber material as the seat. There are also cushioning materials in the front and rear to reduce the impact of the impact. Some materials will be cheaper than the metal plates of traditional cars.
    \item \textit{Interior}: Although the interior of a self-driving car will add a lot of experience products, such as in-car screens, in-car sensors, desks, stereos, etc., but because many traditional car functions will be transferred to the mobile phone APP, The overall cost may still be cheaper than ordinary cars.
\end{itemize}

\paragraph{Operating cost}
Maintenance should be the largest cost in autonomous driving operations. A self-driving car repairers are not as easy to recruit as regular car repairers. Not only must they be rigorously trained, but they also need to have software engineers to assist them in maintenance \cite{bosch2018cost}. Various parts and batteries also need to be replaced from time to time. However, because each driving behavior of an autonomous vehicle can be strictly calculated, it will not step on the accelerator or brake like a human driver, and it can also avoid hard-to-open road sections (such as climbing), and the maintenance of parts is easier.

\paragraph{Social cost}
Big data is the subject of many sectors and is no different for the automotive industry. Nevertheless, the automotive industry will become a collector of data, and a big database developer with the introduction of self-driving vehicles. An autonomous vehicle can generate 100GB of data per second \cite{schlosser2018you}. Self-driving cars depend on huge quantities of data transmitted by various sensors installed. Cars that drive will know exactly where they are and where they are going to track something that is encountered when driving \cite{lafrance2016self}. The world in which the vehicle is situated and those that use it must always be recognized by self-drive vehicles. The better an auto is, the more useful it is for car consumers. However, cars need more and more personal data to make them smart and incorporate the data results into the services provided. 

Users are paying more attention to the threat of AVs to their privacy. Self-driving vehicles are likely to compound this problem. Consumers and regulatory authorities are conscious that vast amounts of personal data and information about consumers and the environment will be produced, used, and preserved in these vehicles. Automobiles are real data factories. It means that anonymity is compromised in the world mobile surveillance ensures that. From the perspective of protecting privacy, the potential data that self-driving cars can collect and how these data are used have received increasing attention.

A travel mode is one of the most important results. Automobiles provide historical geographical data and continuous geographical data in real-time. Such details would not only help a third party to know where the user is and where the user is now, but it will also know where the user is. Announcers should be able to map user-frequented shops to recognize customer purchasing habits. Insurance undertakings may recognize particular personality types \cite{kannadhasanrecent} by tracking users' daily activities or eating habits, such as frequent fast food restaurants or gyms.

The constant information gathering of cars is a cause for worrying that personal information is used for the purpose of targeted ads and advertisements, and drivers may be bored or even dangerous to others. The auto-driving sensors continuously monitor the surroundings and collect photographs would result in potential privacy breaches. The users of auto-driven vehicles are likely to worry about gathering or using their personal details without their consent, intent, or effect on users themselves.

Self-moving vehicles gather and show information about where, when, and how people travel from one location to another, basically automatically. Users are concerned about the reason for which such personal information is to be used. How do we gather this information? What is the purpose of the data? How long is the data maintained? Who is able to access the information? Some analysts suggest that the user adoption of autonomous vehicles would be influenced significantly by safety issues. \textcolor{black}{Anderson et al. \cite{anderson2014autonomous} believe that autonomous driving will lead to additional costs such as more travel, destruction of parking income and loss of work. Litman \cite{litman2017autonomous} has summed up the cost to vehicles and infrastructure of autonomous driving, threatening, anonymity, increased transport, social equity, and lower employment. A summary of AV passenger's benefits and cost are provided in Table \ref{benefit-vs-cost}.}




\begin{table}[htbp]
\caption{Summary of Passenger Benefit (B) and Cost (C)}
\label{benefit-vs-cost}
\begin{center}
\begin{tabular}{p{6.5cm}c}

\hline  
\textbf{Effect} &  \textbf{Passenger}\\
\hline 
Saved time because the driver would do something else in the car & B\\
Saved fuel because of faster driving & B\\
Decreased emissions from forests due to reduced fuel consumption & B\\
Improved safety in traffic & B\\
Enhanced travel due to lower operating costs & B\\
Developing equipment for cars & C\\
Superior maintenance expenses & C\\
Introduced change posts as the travel markets do not apply fixed social risk rates to customers & B/C\\
Modified charges for AVs repair & B/C\\
Seen improvements in AV security and privacy & B/C\\
Infrastructure maintenance prices have increased & B/C\\
Changed land use & B/C\\
Changed congestion & B/C\\
\hline
\end{tabular}
\end{center}
\end{table}


\subsection{Future Research Directions}

\begin {figure*} [htbp]
\centering
\includegraphics [width=.75\textwidth] {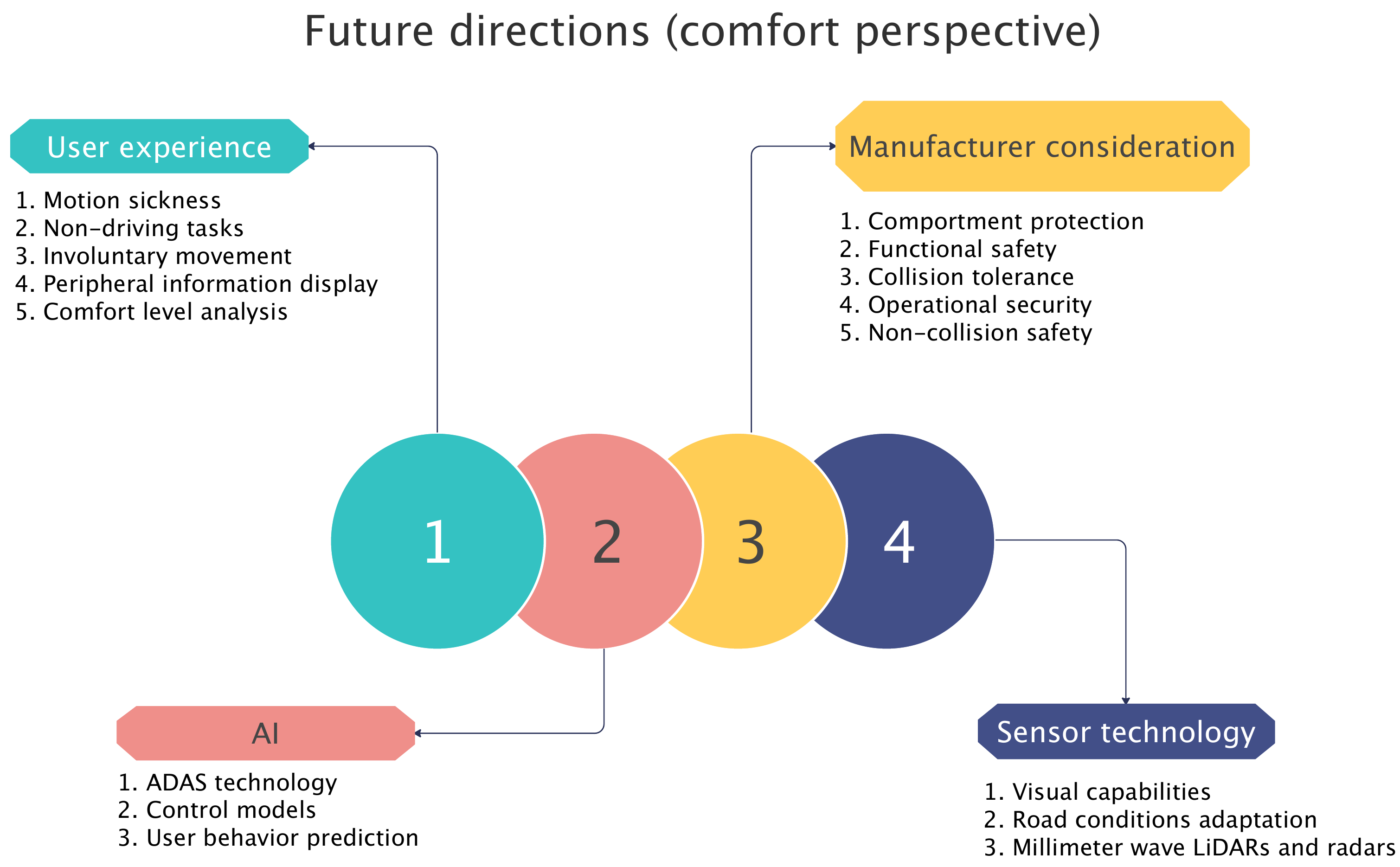}
\caption {Taxonomy of potential future research directions in AV comfort.}
\label{future-dir}
\end {figure*}

Because the technology is not yet mature, and the mass production of autonomous vehicles will take time, autonomous driving still has a long way to go before landing at the L4 and L5 levels. Autonomous driving must confront various problems if it is to be widely used. In addition to advanced sensor technology, more sensor fusion technology is needed. There are still many technical obstacles to be solved. Secondly, there is no doubt about regulation. The surrounding environment in which autonomous vehicles can be used should be clearly specified. In addition, consumers' familiarity and acceptance of autonomous driving solutions is also a point that cannot be ignored. Autonomous driving must allow people to safely give control of driving to a set of very intelligent processors, sensors and software systems behind it. The arrival of driverless driving may take some time, but there will always be a day when it will become popular. In this case, it is not just the type and way people buy cars, but also how they own cars. This section discusses potential directions for future research work focusing on the comfort aspects of AV users. A taxonomy of these directions is depicted in Figure \ref{future-dir} and discussed in detail next.

\subsubsection{Motion Sickness Reduction}
Although motion sickness is an unavoidable problem for autonomous driving (i.e., about 40\% of people have serious symptoms of motion sickness due to severe discomfort), comfortable autonomous driving can support our various activities in the car and reduce motion sickness severity. A future improvement direction can be about addressing the car's stimuli making the vestibule to the center and sense which directly contributes to motion sickness problems. The stimuli here are mainly generated when the vehicle undergoes a linear acceleration (deceleration) speed change, and acceleration and deceleration braking.

The overall comfort of autonomous driving can be enhanced through jerks and accelerations. This can also improve the AV user's experience and enable sustainable driverless services. Therefore, more research work is still needed to investigate the comfort levels in AVs from the perspective of the different jerk and acceleration factors in order to provide proper jerk and acceleration criteria for planning and self-driving strategies. Furthermore, it is highly recommended that future planning and controller solutions for AVs are verified and validated over real-time software environments in accordance with different acceleration criteria.

Last, proposed solutions for motion sickness optimization must be validated based upon a comparison analysis between autonomous driving and how the human driver operates in terms of the driving comfort experienced by both drivers and passengers.

\subsubsection{Enhancement of Non-driving Tasks Experience}
As AV comfort also concerns both the drivers and passengers, non-driving related jobs must be considered and tested in accordance with safety and comfort design practices. These non-driving activities or tasks can be challenging when designing a universal prototype to improve the users' situation awareness. Hence, further automotive research works may focus on specific non-driving tasks inside the AV to maximize the overall comfort experience of the drivers and passengers.

\subsubsection{Modalities of Peripheral Information Display}
To improve the situation awareness of AV users, further research work may consider testing a broad range of modalities to convey peripheral information in the vehicle. Up to date, research works have focused on a single modality of peripheral information display such as the haptic cue. Such a single modality can improve situation awareness, but however, it cannot mitigate motion sickness. Therefore, future research direction could be the exploration and integration of various modalities to enhance the overall awareness of AV users.

\subsubsection{Minimizing Involuntary Movement}
To enhance the AV users' experience and comfort, active mechanisms for movements can be considered for the optimization of AV users' involuntary movement. This can help avoid or reduce the prolonged postural instability that leads to the increase of motion sickness over time. Specifically, reducing or eliminating the involuntary movement of AV users by countering the impacts of centrifugal force upon turning into corners is a potential research direction in improving the overall AV comfort level. This direction will require collaborations between human factors experts and automotive designers and engineers.

\subsubsection{Visual Capabilities and Road Conditions Adaptation}
Improving visual ability of the car is also a challenge faced by current driverless cars. The driverless car needs to recognize other vehicles in the surroundings and must also be able to detect the surrounding lanes, pedestrians, traffic signs, etc. in various challenging environments (such as rain and snow). In addition, challenging road conditions are another problem that driverless cars need to consider and solve.

\subsubsection{AI and Driving Comfort}
The simulation of human intelligence via AI is drastically influencing the growing number of AVs and the sophisticated services they offer. Over 90\% of all car accidents are typically due to driver errors (e.g., poor reaction to the violation or a hazard of traffic laws, poor anticipation, etc.). Hence, further integration of AI techniques in future autonomous driving technology can help with the stagnation in the road accidents rate decline, while increasing the comfort and trust of users in AVs.

Furthermore, governments and legislative authorities may consider making ADAS technology mandatory in all future AVs. ADAS technology will have an AI module to monitor, analyze, and recognize possibly unsafe and dis-comfortable driving behaviors. Specifically, if comfort-disturbing behaviors, such as distractions, drowsiness, and lane departure, are detected, the system can issue a real-time alert warning the driver about the potential danger.

AI-enabled driving control methods can improve the overall driving experience. For instance, navigation systems leverage a broad range of information (e.g., road traffic conditions, real-time weather conditions, etc.) to optimally guide the driver to their destination with the least hassle. A similar component can be integrated as a sub-module of the overall intelligent speed assistance system to support fuel efficiency by providing suggestions about the ideal moments to brake/accelerate to the driver \cite{drewitz2020towards}, \cite{vitale2021caramel}.

\subsubsection{Manufacturer Considerations}

When designing future AV, the key distinct areas of safety and comfort that must be addressed are interpersonal protection, operating safety, crash safety, and health without a crash. These critical factors are summarized as follows.
\begin{itemize}
\item Comportment protection requires driving decisions and actions on the lane. AVs also need to abide by traffic rules, and must provide navigation for users to ensure driving safety in various expected and unexpected scenarios. This problem can be formulated based on precise safety and comfort criteria, where a multi-pronged test over various methods and drive assays must be used.
\item Functional safety helps to guarantee \textcolor{black}{security} tests in AVs in the case of a device malfunction and a backup framework for adapting to the vehicle's unexpected circumstances.
\item Collision tolerance refers to the driver's ability to shield occupants in the car by different steps, shielding people in the driver by structural construction, including seat belts and airbags, and minimizing injuries.
\item Operational \textcolor{black}{security} refers to the interaction between the vehicle and passengers (i.e., HMI). When the safety of operation is ensured, AV can ensure that consumers can bring the comfortable experience provided by AVs. The company will consider enhancing the users' comfort by integrating an interface that allows them to understand their destination, direct vehicles to park by the side, and contact the car's manufacturer to obtain technology auxiliary.
\item Non-collision safety for people who may interact with vehicles. The AVs' manufacturers must aim to provide physical protection against hazards caused by electronic systems or sensors.
\end{itemize}

\subsubsection{Efficient Comfort Level Analysis}
The AV companies must consider employing different methods for safety and risky comfort levels analysis. The risk analysis methods will help define the necessary specifications for autonomous driving system design, subsystems, and components. The designated safety and comfort standards need to follow a set of methods for device and system review, different procedures for systems engineering construction, and the specifications of international and national legislation.


\section{Conclusions} \label{conclusions}
With the promotion of artificial intelligence (AI) technology, self-driving cars are showing fast-paced development, mainly reflected in environmental awareness, decision-making and planning, control and execution, high-precision maps, and real-time positioning technologies. This article explains the comfort of autonomous driving from various perspectives. With that being said, comfort and its associated factors are critical indicators in self-driving cars. Although comfort levels vary from one user to another, they are impacted by many factors, such as the reaction time of autonomous vehicles (AVs), which impacts the comfort of passengers both physiologically and psychologically. Furthermore, when it comes to the planning and design of self-driving cars, as far as the performance of self-driving cars and their user comfort are concerned, user comfort is often ignored. This is typically due to the lack of comprehensive investigations on the control mechanisms in regard to self-driving service operations that relate to user comfort driving factors. This paper discussed the various possibilities impacting the AV's user comfort based on a broad range of technical aspects, biomedical sensors, and psychological analysis. This paper can serve as a baseline for researchers to further address the issues impacting the driver's reaction time in an incident, the level of comfort of the AV, the movement dysfunction, smoothness and jerkiness, and whether autonomous driving is relaxed and safe.

\bibliographystyle{ieeetr}
\bibliography{references.bib}

\begin{IEEEbiography}[{\includegraphics[width=1in,height=1.25in,clip,keepaspectratio]{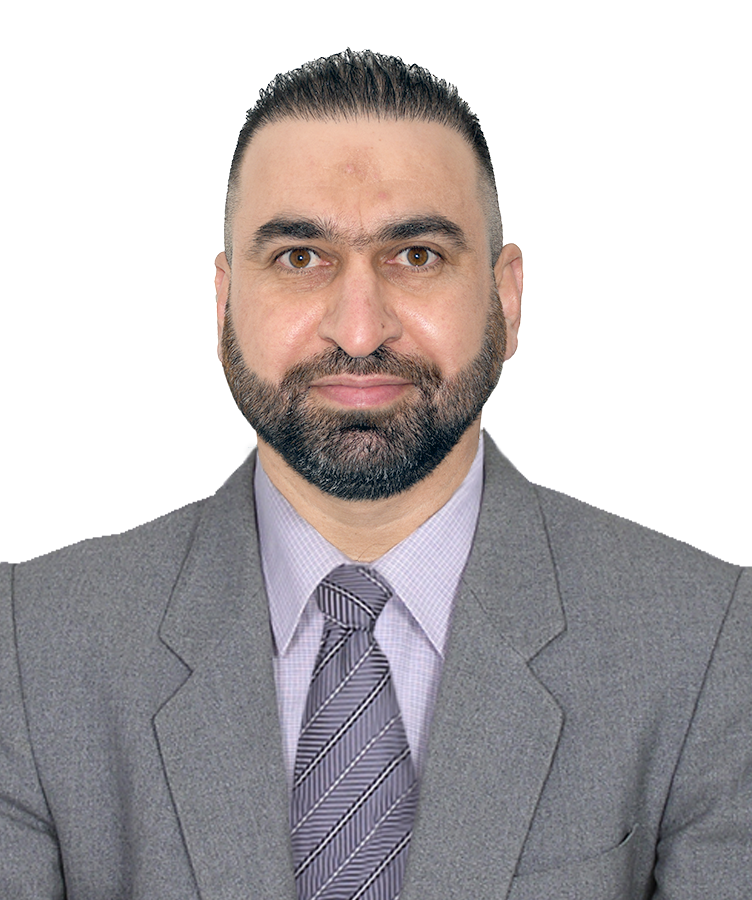}}]{Mohammed Aledhari}(S'14-M'18-SM'20] received his Ph.D. degree in Computer Science from Western Michigan University (WMU), Kalamazoo, Michigan, USA, in January 2018. He is currently an Assistant Professor of Data Science at the University of North Texas (UNT), TX, USA, and is the director of the Computational Healthcare and Biotechnology (COHB) Lab at UNT since September 2022. His research interests include Data Science, Artificial Intelligence, Machine Learning, Computer Vision, Genomics and Computational Biology, Gene and RNA Editing, and Personalized Medicine. Before joining UNT, Dr. Aledhari was an Assistant Professor of Computer Science at Kennesaw State University (KSU), GA, USA. He also served as a Data Scientist at the City of Hope, Clinical Research Center and Hospital before getting his Ph.D. and Postdoc. Dr. Aledhari is a member of IEEE, ACM, and ASQ.
\end{IEEEbiography}

\begin{IEEEbiography}
[{\includegraphics[width=1in,height=1.25in,clip,keepaspectratio]{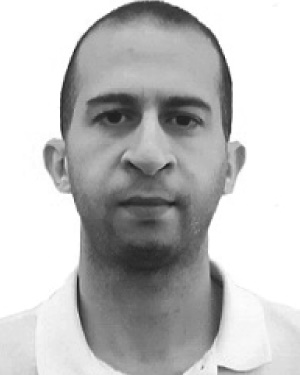}}]
{Mohamed Rahouti} [M'21] received the M.S. degree in Mathematics \& Statistics and the Ph.D. degree in Electrical Engineering from the University of South Florida, Tampa, FL, USA, in 2016 and 2020, respectively. He is currently an Assistant Professor at the Department of Computer and Information Sciences in Fordham University, Bronx, NY, USA. His current research focuses on computer networking and security, blockchain technology, Internet of Things (IoT), machine learning, and network security with applications to smart cities.
\end{IEEEbiography}

\begin{IEEEbiography}
[{\includegraphics[width=1in,height=1.25in,clip,keepaspectratio]{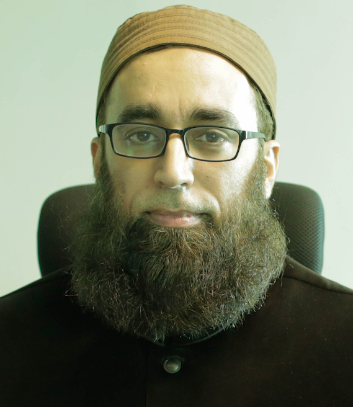}}]
{Junaid Qadir} [SM'14] is a Professor of Computer Engineering at the Qatar University in Doha, Qatar. His primary research interests are in the areas of computer systems and networking, applied machine learning, using ICT for development (ICT4D); human-beneficial artificial intelligence; ethics of technology, artificial intelligence, data science; and engineering education. He has published more than 150 peer-reviewed articles at various high-quality research venues including publications in top international research journals including IEEE Communication Magazine, IEEE Journal on Selected Areas in Communication (JSAC), IEEE Communications Surveys and Tutorials (CST), and IEEE Transactions on Mobile Computing (TMC). He is one of the leading researchers in the area of the security of ML in the context of smart cities, AVs, and networks. He has also given tutorials on the topic of adversarial ML attacks on networks and robust defense at various international conferences including IEEE Vehicular Technology Conference (VTC), Spring-2019 and IWCMC 2019. He was awarded the highest national teaching award in Pakistan—the higher education commission's (HEC) best university teacher award—for the years 2012-2013. He has been appointed as ACM Distinguished Speaker for a three-year term starting in 2020. He is a senior member of IEEE and ACM.
\end{IEEEbiography}


\begin{IEEEbiography}
[{\includegraphics[width=1in,height=1.25in,clip,keepaspectratio]{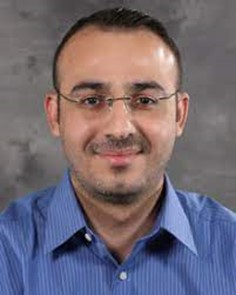}}]{Basheer Qolomany}
[S'17-M'19] received the B.Sc. and M.Sc. degrees in Computer Science from the University of Mosul, Iraq, in 2008 and 2011, respectively, and the Ph.D. and second master's en-route to Ph.D. degrees in Computer Science from Western Michigan University (WMU), Kalamazoo, MI, USA, in 2018. He has worked as a Lecturer with the Department of Computer Science, University of Duhok, Kurdistan region of Iraq, from 2011 to 2013; a Graduate Doctoral Assistant with the Department of Computer Science, WMU, from 2016 to 2018; and a visiting Assistant Professor at the Department of Computer Science, Kennesaw State University (KSU), Marietta, GA, USA, from 2018 to 2019. He is currently an Assistant Professor with the Department of Cyber Systems, University of Nebraska at Kearney (UNK), Kearney, NE, USA. His research interests include evolutionary computation, machine learning, deep learning, and big data analytics in support of population health and smart services. He is a member of IEEE and ACM. 

\end{IEEEbiography}
\begin{IEEEbiography}
[{\includegraphics[width=1in,height=1.25in,clip,keepaspectratio]{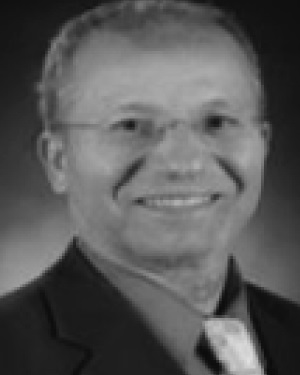}}]
{Mohsen Guizani} (S'85–M'89–SM'99–F'09) received the BS (with distinction), MS and PhD degrees in Electrical and Computer engineering from Syracuse University, Syracuse, NY, USA in 1985, 1987 and 1990, respectively. He is currently a Professor of Machine Learning and the Associate Provost at Mohamed Bin Zayed University of Artificial Intelligence (MBZUAI), Abu Dhabi, UAE. Previously, he worked in different institutions in the USA. His research interests include applied machine learning and artificial intelligence, Internet of Things (IoT), intelligent autonomous systems, smart city, and cybersecurity. He was elevated to the IEEE Fellow in 2009 and was listed as a Clarivate Analytics Highly Cited Researcher in Computer Science in 2019, 2020, 2021 and 2022. Dr. Guizani has won several research awards including the “2015 IEEE Communications Society Best Survey Paper Award”, the Best ComSoc Journal Paper Award in 2021 as well five Best Paper Awards from ICC and Globecom Conferences. He is the author of ten books and more than 800 publications. He is also the recipient of the 2017 IEEE Communications Society Wireless Technical Committee (WTC) Recognition Award, the 2018 AdHoc Technical Committee Recognition Award, and the 2019 IEEE Communications and Information Security Technical Recognition (CISTC) Award. He served as the Editor-in-Chief of IEEE Network and is currently serving on the Editorial Boards of many IEEE Transactions and Magazines. He was the Chair of the IEEE Communications Society Wireless Technical Committee and the Chair of the TAOS Technical Committee. He served as the IEEE Computer Society Distinguished Speaker and is currently the IEEE ComSoc Distinguished Lecturer.

\end{IEEEbiography}

\begin{IEEEbiography}
[{\includegraphics[width=1in,height=1.25in,clip,keepaspectratio]{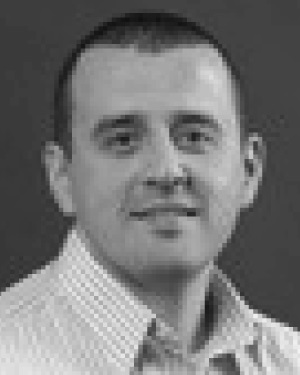}}]
{Ala Al-Fuqaha} [S'00-M'04-SM'09] received Ph.D. degree in Computer Engineering and Networking from the University of Missouri-Kansas City, Kansas City, MO, USA, in 2004. He is currently Professor and Associate Provost at Hamad Bin Khalifa University (HBKU). His research interests include the use of machine learning in general and deep learning in particular in support of the data-driven and self-driven management of large-scale deployments of IoT and smart city infrastructure and services, Wireless Vehicular Networks (VANETs), cooperation and spectrum access etiquette in cognitive radio networks, and management and planning of software defined networks (SDN). He is a senior member of the IEEE and an ABET commissioner and Program Evaluator (PEV). He serves on editorial boards of multiple journals including IEEE Communications Letter and IEEE Network Magazine. He also served as chair, co-chair, and technical program committee member of multiple international conferences including IEEE VTC, IEEE Globecom, IEEE ICC, and IWCMC.
\end{IEEEbiography}

\end{document}